%% file: main.tex
\documentclass{article}

\usepackage{arxiv}

\usepackage{wrapfig}
\usepackage{booktabs} 
\usepackage{caption}
\usepackage{subcaption}
\usepackage{amsmath,amssymb,amsfonts}
\usepackage{algorithmic}
\usepackage[numbers]{natbib}
\usepackage{comment}
\usepackage[linesnumbered,ruled,vlined]{algorithm2e}
\SetKwComment{Comment}{$\triangleright$\ }{}

\usepackage{graphicx}
\usepackage{textcomp}
\usepackage{multirow}
\usepackage{hyperref}
\usepackage[table, svgnames, dvipsnames]{xcolor}
\usepackage{makecell, cellspace, caption}
\def\BibTeX{{\rm B\kern-.05em{\sc i\kern-.025em b}\kern-.08em
    T\kern-.1667em\lower.7ex\hbox{E}\kern-.125emX}}

\definecolor{songcolor}{RGB}{46,139,87}

\def\A{{\bf A}}
\def\a{{\bf a}}

\def\b{{\bf b}}

\def\c{{\bf c}}
\def\D{{\bf D}}

\def\h{{\bf h}}

\def\H{{\bf H}}

\def\M{{\bf M}}

\def\U{{\bf U}}

\def\v{{\bf v}}
\def\W{{\bf W}}
\def\w{{\bf w}}
\def\X{{\bf X}}
\def\x{{\bf x}}

\def\y{{\bf y}}

\def\0{{\bf 0}}
\def\1{{\bf 1}}
\def\RB{{\mathbb R}}

\def\maxpool{\mathsf{MaxPool}}

\title{Graph Message Passing with Cross-location Attentions for Long-term ILI Prediction} 

\author{
Songgaojun Deng,\textsuperscript{1} Shusen Wang,\textsuperscript{1} Huzefa Rangwala,\textsuperscript{2} Lijing Wang,\textsuperscript{3} Yue Ning\textsuperscript{1}\\
\textsuperscript{1}{Stevens Institute of Technology}\\
\textsuperscript{2}{George Mason University}\\
\textsuperscript{3}{University of Virginia}\\
sdeng4, shusen.wang, yue.ning@stevens.edu, rangwala@cs.gmu.edu, lw8bn@virginia.edu
}

\begin{document}
\maketitle
\begin{abstract}
Forecasting influenza-like illness (ILI)  is of prime importance to epidemiologists and health-care providers. Early prediction of epidemic outbreaks plays a pivotal role in disease intervention and control. Most existing work has either limited long-term prediction performance or lacks a comprehensive ability to capture spatio-temporal dependencies in data.
Accurate and early disease forecasting models would markedly improve both epidemic prevention and managing the onset of an epidemic.
In this paper, we design a \underline{\textbf{c}}r\underline{\textbf{o}}ss-\underline{\textbf{l}}ocation \underline{\textbf{a}}ttention based \underline{\textbf{g}}raph \underline{\textbf{n}}eural \underline{\textbf{n}}etwork (Cola-GNN) for learning time series embeddings and location aware attentions. We propose a graph message passing framework to combine learned feature embeddings and an attention matrix to model disease propagation over time. We compare the proposed method with state-of-the-art statistical approaches and deep learning models on real-world epidemic-related datasets from United States and Japan. The proposed method shows strong predictive performance and leads to interpretable results for long-term epidemic predictions.
\end{abstract}

\input{sections/introduction.tex}
\input{sections/relatedwork.tex}
\input{sections/method.tex}
\input{sections/experiment.tex}
\input{sections/result.tex}

\section{Conclusion}
In this work, we propose a graph-based deep learning framework with cross-location attentions to study the spatio-temporal influence of long-term epidemiological predictions. We demonstrate the effectiveness of the proposed model on real-world epidemiological datasets. The proposed method is not flexible enough in the case that different models are trained for different lead time settings. Future work will consider iterative predictions to increase the flexibility of the model. Another research direction is to involve more complex dependencies such as weather, social factors, and population migration. We intend to determine if the prediction accuracy is improved when using external indicators. Furthermore, it is also essential to identify the main factors affecting the epidemic outbreak of one area by learning multiple areas simultaneously.

\begin{footnotesize}
\bibliographystyle{plainnat}
\bibliography{ref}
\end{footnotesize}%

\end{document}

%% file: sections/introduction.tex
\section{Introduction}

Epidemic disease propagation that involves large populations and wide areas can have a significant impact on society. The Center for Disease Control and Prevention (CDC) estimates 79,400 deaths from influenza occurred during the 2017-2018 season in the United States~\footnote{\url{https://tinyurl.com/y3tf8ebl}}.
Early forecasting of infectious diseases such as influenza-like illness (ILI) provides optimal opportunities for timely intervention and resource allocation.
It helps with the timely preparation of corresponding vaccines in health care departments which leads to reduced financial burden.
For instance, 
the World Health Organization (WHO) reports that  Australia spent over 352 million dollars on routine immunization in the 2017 fiscal year~\footnote{\url{https://tinyurl.com/y2duz5p8}}.
We focus on the problem of long term ILI forecasting with lead time from 1 to 20 weeks based on the influenza surveillance data collected for multiple locations (states and regions).
Given the process of data collection and surveillance lag, accurate statistics for influenza warning systems are often delayed by a few weeks, making early prediction imperative. However, there are a few challenges in long-term epidemic forecasting. First, the temporal dependency is hard to capture with short-term input data. Without manually added seasonal trends, most statistical models fail to provide high accuracy. Second, the influence from other locations has not been exhaustively explored with limited data input. Spatio-temporal effects have been studied but they usually require adequate data sources to achieve good performance~\cite{senanayake2016predicting}. 

Existing work on epidemic prediction has been focused on various aspects: 
1)  Traditional causal models~\cite{kermack1927contribution,chowell2008seasonal,bisset2009epifast}, including compartmental models and agent-based models, employ disease progression mechanisms such as Susceptible-Infectious-Recovered (SIR) to capture the dynamics of ILI diseases. Compartmental models focus on mathematical modeling of population-level dynamics. Agent-based models simulate the propagation process at the individual level with contact networks. Calibrating these models is challenging due to the high dimensionality of the parameter space. 
 2) Time series prediction with statistical models such as Autoregressive (AR) and its variants (e.g., VAR) are not suitable for long term ILI trend forecasting given that the disease activities and human environments evolve over time. 
 3) Machine learning and deep learning methods~\cite{santos2014analysing,venna2018novel,wu2018deep,Wang-2019} such as recurrent neural networks have been explored in recent years but they barely consider cross-spatial effects in long term disease propagation.


In this paper, we focus on long term (10-20 weeks) prediction of the count of ILI patients using data from a limited time range (20 weeks). To tackle this problem, we explore a graph propagation model with deep spatial representations to compensate the loss of temporal information. Assuming each location is a node, we design a graph neural network framework to model epidemic propagation at the population level. Meanwhile, we investigate recurrent neural networks for capturing sequential dependencies in local time series data and temporal convolutions for identifying short-window patterns. Our key contributions are summarized as follows:

\begin{itemize}
    \item We propose a novel graph-based deep learning framework for long-term epidemic prediction from a time-series forecasting perspective. 
  This is one of the first works of graph neural networks adapted to epidemic forecasting.
    
    \item We investigate a location-aware attention mechanism to capture location correlations. The influence of locations can be directed and automatically optimized in the model learning process. The attention matrix is further evaluated as an adjacency matrix in the graph neural network for modeling disease propagation.
    
    \item We design a temporal convolution module to automatically extract temporal dependencies and hidden features for time series data of multiple locations. The learned temporal features for each location are utilized as node attributes for the graph neural network.
    
    \item The proposed method, Cola-GNN, outperforms a broad range of state-of-the-art models on three real-word datasets with different long-term prediction settings. We also demonstrate the effectiveness of its learned attention matrix compared to a geographical adjacency matrix in an ablation study.
\end{itemize}


%% file: sections/relatedwork.tex
\section{Related Work}

\subsection{Influenza Prediction}

In many studies, forecasting influenza or influenza-like illnesses (ILI) case counts is formulated as time series regression problems, where autoregressive models are widely used~\cite{viboud2003prediction,achrekar2011predicting,dugas2013influenza,wang2015dynamic}. 
Instead of focusing on seasonal effects, \citeauthor{wang2015dynamic} \cite{wang2015dynamic} propose a dynamic poisson autoregressive model to improve short-term prediction accuracy (e.g. 1-4 weeks). 
Furthermore, variations of particle filters and ensemble filters have been used to predict influenza activities. 
\citeauthor{yang2014comparison} \cite{yang2014comparison} evaluate the performance of six state-of-the-art filters to forecast influenza activity and concluded that the models have comparable performance. 
Ensemble methods such as matrix factorization based regression and nearest neighbor based regression have been studied~\cite{chakraborty2014forecasting}. 
While autoregressive, filter-based, and ensemble models are simple and straightforward, they often neglect the geographical dependence in disease propagation.

Attempts to study spatio-temporal effects in influenza disease modeling are not rare.
\citeauthor{waller1997hierarchical} propose a hierarchical Bayesian parametric model for the spatio-temporal interaction of generic disease mapping~\cite{waller1997hierarchical}. 
A non-parametric Bayesian method~\cite{senanayake2016predicting} is proposed for predicting spatial and temporal variation of influenza cases.
\citeauthor{venna2018novel} develop data-driven approaches involving climatic and geographical factors for real-time influenza forecasting~\cite{venna2018novel}.  
\citeauthor{wu2018deep} use deep learning for modeling spatio-temporal patterns in epidemiological prediction problems~\cite{wu2018deep}.
Despite their impressive performance, these methods have limitations such as the requirement of additional data which are not readily available and long-term prediction performance is not satisfactory.
Improving the long-term epidemiological prediction with restricted training data is an open research problem.


\subsection{Long-term Epidemic Prediction}

Long-term prediction (aka multi-step prediction), that is, predicting several steps ahead, is a challenge in time series prediction.
Long-term prediction has to face growing uncertainties arising from various problems such as accumulation of errors and lack of information.
Long-term prediction methods can be categorized into two types: (i) direct methods and (ii) iterative methods.
Direct methods predict a future value using the past values in one shot.
Iterative methods recursively invoke short-term predictors to make long-term predictions.
Specifically, they use the observed data $x_1, \dots, x_t$ to predict the next step $x_{t+10}$, then use $x_2, \cdots, x_{t+1}$ to predict $x_{t+11}$, and so on.

For long-term predictions using time series data,
\citeauthor{sorjamaa2007methodology} combine a direct prediction strategy and sophisticated input selection criteria~\cite{sorjamaa2007methodology}; 
\citeauthor{qian2008multi} and \citeauthor{du2014prediction} develop neural network based methods to improve the performance of long-term prediction~\cite{qian2008multi,du2014prediction}.
Recent works~\cite{venna2018novel,wu2018deep} explore deep learning models for direct long-term epidemiological predictions.
DEFSI~\cite{Wang-2019} combines deep neural network methods with causal models to address high-resolution ILI incidence forecasting. Yet most of these models rely heavily on extrinsic data to improve accuracy.

%% file: sections/method.tex
\section{The Proposed Method}

\subsection{Problem Formulation}

We formulate the epidemic prediction problem as a regression task with multiple time series as input.
Throughout the paper, we denote the number of locations by $N$ and the time span for one input example as $T$. We use the terms region and location interchangeably.

At each time step $t$, the multi-location epidemiology profile is denoted by $\mathbf{x}_t \in \mathbb{R}^N$ whose elements are the observations from $N$ sources/locations, e.g. the influenza patient counts per week $(t)$ in $N$ locations. We further denote the training data in a time-span of size $T$ as $\mathbf{X}=[\mathbf{x}_1,...,\mathbf{x}_T] \in \mathbb{R}^{N\times T}$. The objective is to predict an epidemiology profile at a future time point $T+h$ where $h$ refers to the horizon/lead time of the prediction. 

\begin{figure}
 \centering
 \includegraphics[width=0.8\textwidth]{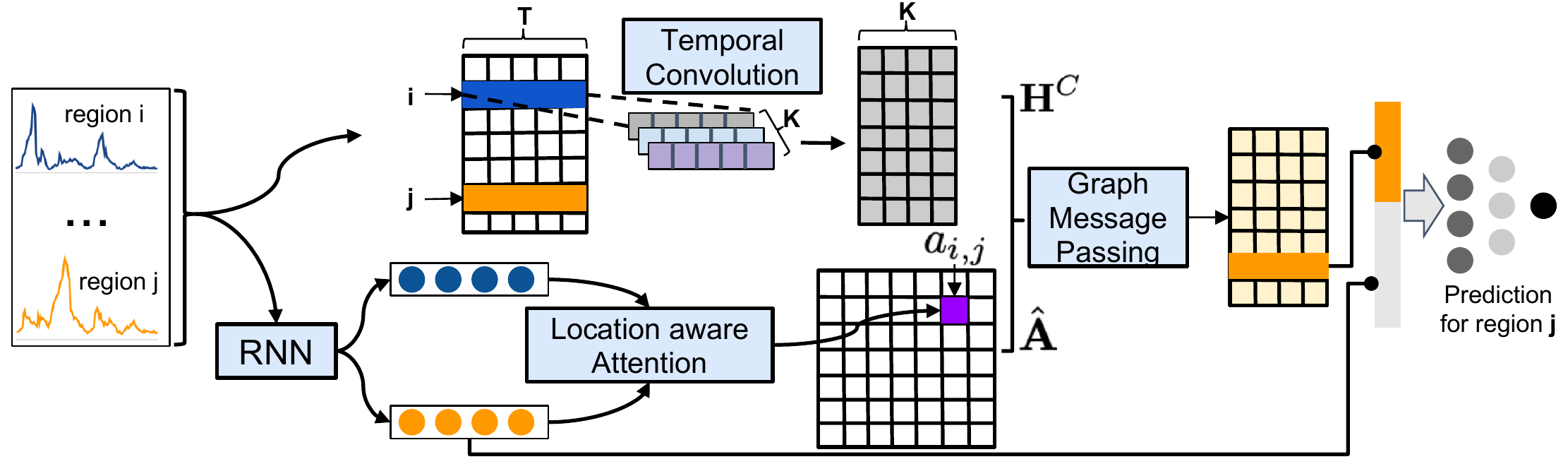} 
  \caption{The overview of the proposed framework. Eq.~\ref{eq:ag-norm}-\ref{eq:attn-fusion} are skipped for brevity.}
  \label{fig:model}
\end{figure}

The proposed 
framework as shown in Figure~\ref{fig:model} consists of three modules: 1) location-aware attention to capture location wise interactions, 2) temporal convolutional layer to capture local temporal features,  3) global graph message passing to combine the temporal features and the location-aware attentions to generate further hidden features and make predictions. The pseudocode is described in Algorithm~\ref{alg:model} and each module is described as below. 
\begin{algorithm}
\DontPrintSemicolon
\KwIn{Time series data $\{\X,\y\}$ from multiple locations, geographical  adjacency  matrix $\A^{g}$}
\KwOut{Model parameters $\Theta$}
\For{each epoch}{
    Randomly sample a mini batch\;
    \For{each region $i$}{
    $\h_{i,T} \gets \textup{RNN module}(\x_{i:})$\;
    $\h_{i}^{C} \gets \textup{Temporal Conv}(\x_{i:})$\;
    }
    \For{each region pair $(i,j)$}{
    $\hat{a}_{i,j} \gets \textup{Loc-Aware Attn}(\h_{i,T},\h_{j,T},\A^{g})$\;
    }
    \Comment*[r]{\textup{Simultaneous calculations for all regions}}
    \For{each region $i$}{
    $\h_{i}^{l} \gets \textup{Graph Message Passing}(\h_{i}^{C},\hat{\A})$\;
     $\hat{y}_i \gets \textup{Output}\big([\h_{i,T} ; \h_i^{(l)}]\big)$\;
    }
     $\Delta \mathcal{L}(\Theta) \gets \textup{BackProp}\big(\mathcal{L}(\Theta), \y,\hat{\y}, \Theta\big)$\;

    $\Theta \gets \Theta-\eta \Delta\mathcal{L}(\Theta)$ \Comment*[r]{\textup{SGD step}}
}
\caption{\textbf{Cola-GNN}}
\label{alg:model}
\end{algorithm}

\subsection{Location-aware Attention}\label{sec:attention}
In this study, without precise population movement data, we dynamically model the impact of one area on other areas during the epidemics of infectious disease. 
We first learn hidden states for each location given a time period using a
Recurrent Neural Network (RNN) given its great success in sequential (temporal) data prediction. 
Specifically, we use a simple and classic vanilla RNN  in this module. 
The RNN module can be replaced by Gated Recurrent Unit (GRU)~\cite{cho2014learning} or Long short-term memory (LSTM)~\cite{hochreiter1997long}; however, in this application, RNN achieves the best performance compared to GRU and LSTM. 

Given the  multi-location time series data $\mathbf{X}=[\mathbf{x}_1,...,\mathbf{x}_T] \in \mathbb{R}^{N\times T}$, we employ a global RNN model to capture the temporal dependencies of all locations. For location $i$, an instance of a time series is represented by $\x_{i:}=[x_{i,1},...,x_{i,T}] \in \mathbb{R}^{1\times T}$.
Let $D$ be the dimension of the hidden state.
For each element $x_{i,t}$ in the input sequence, the RNN updates its hidden state according to
\begin{equation}
    \mathbf{h}_{i,t} =\tanh \big(\w x_{i,t}  + \U \h_{i,t-1} + \b \big)
    \in \mathbb{R}^{D},
\label{equ:for}
\end{equation}
where $\h_{i, t}$ is the hidden state vector at time $t$ and $\h_{i,t-1}$ is the hidden state vector at time $t-1$; $\tanh$ is the non-linear activation function; $\w \in \RB^{D}$, $\U \in \RB^{D\times D}$, and $\b  \in \RB^{D}$ 
determine the adaptive weight and bias vectors of the RNN. 
Let $\h_{i} = \h_{i,T}$ be the last hidden state and
we will use it to represent location $i$.




Next, we define an attention coefficient $a_{i,j}$ for measuring the impact of location $j$ on location $i$. 

Additive attention (or multi-layer perceptron attention)~\cite{bahdanau2014neural} and multiplicative attention (or dot-product attention)~\cite{vaswani2017attention,sukhbaatar2015end} are the two most commonly used attention mechanisms. 
They share the same idea of computing the alignment score between elements from two sources, but with different compatibility functions.
We utilize the compatibility function of additive attention due to its better predictive quality, which is defined as:
\begin{equation}
    a_{i,j} =\v^T g (\W^{s}\h_{i}+ \W^{t}\h_{j}+ \b^{s}) + b^v ,
\label{equ:loc-attn-1}
\end{equation}
where $g$ is an activation function that is applied element-wise;
$\W^{s}, \W^{t} \in \RB^{d_{a} \times D}$, $\v \in \RB^{d_{a}}$, $\b^s \in \RB^{d_{a}}$, and $b^v \in \mathbb{R}$ are trainable parameters. $d_{a}$ is a hyperparameter that controls the dimensions of the parameters in Eq.~\ref{equ:loc-attn-1}.
Assuming that the impact of location $i$ on location $j$ is different than vice versa, we obtain an asymmetric attention coefficient matrix $\A$ where each row indicates the degree of influence by other locations on the current location. Usually, a softmax function is used to transform the attention scores to a probability distribution. In our problem, the overall impact of other locations vary for different places. For instance, compared to New York, Hawaii may be less affected overall by other states. 
Instead, we perform normalization over the rows of $\A$ to normalize the impact of other locations on one location:
\begin{equation}
    \a_{i:} \longleftarrow \frac{\a_{i:}  }{\max(\left \| \a_{i:}  \right \|_{p},\epsilon)} ,
\label{equ:norm}
 \end{equation}
where $\epsilon$ is a small value to avoid division by zero, and $\| \cdot \|_p$ denotes the $\ell_p$-norm. 

Given the geographic nature of this task, we also consider the spatial distance between two locations. 
We use $\A^{g}$ to indicate the connectivity of locations: $a^{g}_{i,j}=1$ means locations $i$ and $j$ are neighbors~\footnote{By default, each location is adjacent to itself.}. 
The correlation of the two locations may be affected by their geographic distance, i.e. nearby areas may have similar topographic or climatic characteristics that make them have similar flu outbreaks. Non-adjacent areas may also have potential dependencies due to population movements and similar geographical features. Simulating all the factors related to a flu outbreak is difficult. 
Therefore, we consider both the attention derived from historical data and the geographical distances of the locations. The final location-aware attention matrix is obtained by combining the geographical adjacency matrix $\tilde{\A}^{g}$ and the attention matrix $\A$. The combination is accomplished by an element-wise gate $\M$, learned from the attention matrix which evolves over time. 
We consider the attention matrix to be a feature matrix with gate $\M$ being adapted from the feature fusion gate~\cite{gong2017ruminating}:
\begin{eqnarray}
    \tilde{\A}^{g} & = & \D^{-\frac{1}{2}}\A^{g}  \D^{-\frac{1}{2}}  , \label{eq:ag-norm} \\
    \M & = & \sigma (\W^{m} \A +  b^{m} \1_N \1_N^T ) , \label{eq:attn-gate} \\
    \hat{\A} & = &  \M \odot \tilde{\A}^{g}  +  (\1_N \1_N^T - \M) \odot \A  , \label{eq:attn-fusion}
\end{eqnarray}
where Eq.~\ref{eq:ag-norm} is for normalization, $\D$ is the degree matrix defined as $d_{ii} = \sum_{j=1}^N a^g_{ij}$.   
$\W^m \in \RB^{N \times N}$ and $b^m \in \RB$ are trainable parameters.



\subsection{Temporal Convolution Layer}\label{sec:cnn}
Besides the spatial dependencies,
the outbreak of influenza also has its unique characteristics over time. For instance, the United States experiences annual epidemics of seasonal flu. Most of the time flu activity peaks between December and February, and it can last as late as May~\footnote{\url{https://tinyurl.com/yxevpqs9}}.
Convolutional Neural Networks (CNN) have shown successful results in capturing various important local patterns from grid data and sequence data. 
We apply 1D CNN filters to every row of $\X$ to capture the temporal dependency; note that the row $\x_{s:}$ is the observed sequential data at location $s$.
Specifically, we define $K$ filters where each filter $\c_k \in  \mathbb{R}^{1 \times Q} $ and $Q$ is chosen to be the maximum window length $T$ in our experiments. Convolutional operations yield $\H^C \in \mathbb{R}^{N \times K} $, where $h^C_{i,k}$
 represents the convolutional value of the $i$-th row vector and the $k$-th filter. Formally, this convolution operation is given by
\begin{equation}
h^C_{i,k}=\text{ReLU} \left(\maxpool \Big( \sum_{\tau=1}^{Q}x_{i,\tau} \times c_{k,\tau}\Big) \right) .
\label{equ:temporal-conv}
\end{equation}
Max pooling is needed when $Q<T$ as in \citeauthor{kim-2014-convolutional} \cite{kim-2014-convolutional}.
To constrain the data, we also apply a nonlinearity to the convolution results.
Then the new detected temporal feature of each row/location is $\h^C_{i} = [h^C_{i,1},...,h^C_{i,K}] \in \mathbb{R}^{1 \times K}$. 

\subsection{Graph Message Passing (the Propagation Model)}
After learning the cross-location attentions (Section~\ref{sec:attention}) and the local hidden features (Section~\ref{sec:cnn}), we design a flu propagation model using graph neural networks.
Graph neural networks iteratively update the node features from their neighbors. When generalized to irregular domains, this operation is often referred to as message passing or neighbor aggregation.
Epidemic disease propagation at the population level is usually affected by human connectivity and transmission. Considering each location as a node in a graph, we take advantage of graph neural networks to model the epidemic disease propagation among different locations.
We model the adjacency matrix using the cross-location attention matrix and the nodes' initial features using the the temporal convolutional features.
With $\mathbf{h}_i^{(l-1)} \in \mathbb{R}^{F^{(l-1)}}$ denoting node features of node $i$ in layer $(l-1)$ and $\hat{a}_{i,j}$ denoting the location-aware attention coefficient from node $j$ to node $i$, the message passing graph neural network can be described as  

\begin{equation}
    \mathbf{h}_i^{(l)} = g \Big( \sum_{j \in \mathcal{N}} \hat{a}_{i,j} \mathbf{W}^{(l-1)} \mathbf{h}_{j}^{(l-1)} 
    +\mathbf{b}^{(l-1)}\Big) ,
\label{equ:message-passing}
\end{equation}
where $g$ denotes a nonlinear activation function, 
$\mathbf{W}^{(l-1)} \in \mathbb{R}^{F^{(l)} \times F^{(l-1)}}$ is the weight matrix for hidden layer $l$ with with $F^{(l)}$ feature maps, and $\mathbf{b}^{(l-1)} \in \mathbb{R}^{F^{(l)}}$ is a bias. $\mathcal{N}$ is the set of locations.
$\mathbf{h}_i^{(0)}$ is initialized with $\mathbf{h}_{i}^{C}$ at the first layer.

\subsection{Output Layer (Prediction)}
For each location, we learn the RNN hidden states ($\mathbf{h}_{i,T} \in \mathbb{R}^D$) from its own historical sequence data, as well as the graph features ($\mathbf{h}_i^{(l)} \in \mathbb{R}^{F^{(l)}}$) learned from other locations' data in our propagation model. We combine these two features and feed them to the output layer for prediction, which is defined as:
\begin{equation}
    \hat{y}_i =  \phi \Big( \boldsymbol{\theta}^{\top} [\mathbf{h}_{i,T} ; \mathbf{h}_i^{(l)}] + b^{\theta} \Big) ,
\label{equ:out}
\end{equation}
where $\phi$ is the activation function (identity or nonlinear) and $\boldsymbol{\theta} \in \RB^{D+F^{(l)}}, b^{\theta} \in \mathbb{R}$ are model parameters.
\subsection{Optimization} We compare the prediction value of each location with the corresponding ground truth and then optimize a regularized $\ell_1$-norm loss:
\begin{equation}
    \mathcal{L}(\Theta)= \sum_{i=1}^{N}\sum_{m=1}^{n_i}|y_{i,m}-\hat{y}_{i,m}|+\lambda \mathcal{R}(\Theta) ,
\label{equ:opt}
\end{equation}
where $n_i$ is the number of samples in location $i$ obtained by a moving window, shared by all locations, $y_{i,m}$ is the true value of location $i$ in sample $m$, and $\hat{y}_{i,m}$ is the model prediction. $\Theta$ stands for all training parameters and $\mathcal{R}(\Theta)$ is the regularization term (e.g. $\ell_2$-norm). All model parameters can be trained via back-propagation and optimized by the Adam algorithm~\cite{kingma2014adam} given its efficiency and ability to avoid overfitting.

%% file: sections/experiment.tex
\section{Experiment Setup}
\subsection{Datasets}
We prepare three real-world datasets for experiments: Japan-Prefectures, US-States and US-Regions and their data statistics are shown in Table~\ref{tab:data}.

\begin{itemize}
    \item \textbf{Japan-Prefectures} We collect this data from the Infectious Diseases Weekly Report (IDWR)~\footnote{\url{https://tinyurl.com/y5dt7stm}} in Japan. This dataset contains weekly influenza-like-illness statistics (patient counts) from 47 prefectures in Japan, ranging from August 2012 to March 2019. 
    \item \textbf{US-States} We collect the influenza disease data from the Center for Disease Control (CDC)~\footnote{\label{cdcdata}\url{https://tinyurl.com/y39tog3h}}. 
It contains the count of patient visits for ILI for each week and each state in United States from 2010 to 2017.
After removing a state with missing data we kept 49 states remaining in this dataset.
\item \textbf{US-Regions} This dataset is the ILINet portion of the US-HHS (Department of Health and Human Services) dataset
\addtocounter{footnote}{-1}
~\footnotemark, consisting of weekly influenza activity levels for 10 HHS regions of U.S. mainland for the period of 2002 to 2017. Each HHS region represents some collection of associated states. 
We use flu patient counts for each region, which is calculated by combining state-specific data.
\end{itemize}

Data is normalized to 0-1 range for each region. The maximum value of the region is set to 1, and the minimum value of the region is set to 0. After ordering the data by time, the first 50\% is used for training, next 20\% for validation, and the last 30\% for testing. Validation data is used to determine the number of epochs that should be run to avoid overfitting. We fixed the validation and test sets by dates for different lead time values. In this case, the test data covers 2.1, 4.5, and 2.1 flu seasons in Japan-Prefectures, US-States and US-Regions respectively. Accordingly, there are at least 3, 7.2 and 3 flu seasons in the three training sets. All data is normalized based on the maximum and minimum values of the training data.


\begin{table}
\centering
\caption{Dataset statistics: min, max, mean, and standard deviation (SD) of patient counts; dataset size means number of locations multiplied by \# of weeks. }\label{tab:data}
\begin{tabular}{lccccc}
\toprule
\textbf{Data set} &\textbf{Size}  &\textbf{Min} &\textbf{Max}   &\textbf{Mean}  &\textbf{SD}  \\
\midrule
Japan-Prefectures &47$\times$348 &0 &26635 &655 &1711
\\
US-Regions &10$\times$785 &0 &16526 &1009 &1351
\\
US-States &49$\times$360 &0 &9716 &223 &428
\\
\bottomrule
\end{tabular}
\label{space}
\end{table}

\subsection{Evaluation Metrics}
In the experiments, we adopt the following metrics for evaluation. Denote the prediction and true values to be $\{ \hat{y}_{1},...,\hat{y}_{n} \}$ and $\{y_{1},...,y_{n}\}$, respectively. We do not distinguish regions in evaluation.

The \textbf{Root Mean Squared Error (RMSE)} measures the difference between predicted and true values after projecting the normalized values into the real range:
    $$\text{RMSE} ={\sqrt {\frac {1}{n}\sum _{i=1}^{n}({\hat {y}}_{i}-y_{i})^{2}}}.$$
    
The \textbf{Mean Absolute Error (MAE)} is a measure of difference between two continuous variables:
 $$\text{MAE} ={\frac {1}{n}\sum _{i=1}^{n}\left|\hat{y_{i}}-y_{i}\right|}.$$

The \textbf{Pearson’s Correlation (PCC)} is a measure of the linear dependence between two variables:
$$\text{PCC}={\frac {\sum _{i=1}^{n}(\hat{y}_{i}-{\bar {\hat{y}}})(y_{i}-{\bar {y}})}{{\sqrt {\sum _{i=1}^{n}(\hat{y}_{i}-{\bar {\hat{y}}})^{2}}}{\sqrt {\sum _{i=1}^{n}(y_{i}-{\bar {y}})^{2}}}}}.$$

\textbf{Leadtime} is the number of weeks that the model predicts in advance. For instance, if we use $X_{N,T}$ as input and predict the infected patients of the fifth week (leadtime = 5) after current week $T$, the ground truth (expected output) is $X_{N, T+5}$.

\subsection{Comparison Methods}
We compare our model with several state-of-the-art methods and their variants listed as below.

\begin{itemize}
\item  \textbf{Autoregressive (AR)} Autoregressive models have been widely applied for time series forecasting~\cite{brownstein2017combining,wang2015dynamic}. Basically, the future state is modeled as a linear combination of past data points. We train an autoregressive model for each location. No data and parameters are shared among locations.

\item
\textbf{Global Autoregression (GAR)} This model is mainly used when training data is limited. We train one global model using the data available from each location.
\item 
\textbf{Vector Autoregression (VAR)} The VAR models cross-signal dependence to address the potential drawback of the AR model, i.e. the signal sources are processed independently of each other. Therefore, it introduces more parameters and is more expensive in training.
\item 
\textbf{Autoregressive Moving Average (ARMA)} ARMA contains the autoregressive terms and moving-average terms together. A considerable amount of preprocessing has to be performed before such model fitting. The order of the moving average is set to 2 in implementation.
\item
\textbf{Recurrent Neural Network (RNN)} RNNs have demonstrated powerful abilities to predict temporal dependencies. We employ a global RNN for our problem, that is, parameters are shared across different regions. RNN can be be replaced by GRU or LSTM. Experimentally, fancy RNN models did not achieve better results, so we only consider simple RNN for comparison.
\item 
\textbf{RNN+Attn~\cite{cheng2016long}} This model considers the self-attention mechanism in a global RNN.  
In the calculation of rnn units, the hidden state is replaced by a summary vector, which uses the attention mechanism to aggregate all the information of the previous hidden state.

\item 
\textbf{CNNRNN-Res~\cite{wu2018deep}} A deep learning framework that combines CNN, RNN and residual links to solve epidemiological prediction problems.
\item 
\textbf{GCNRNN-Res} A variation of CNNRNN-Res. We change the CNN module to a GCN~\cite{kipf2016semi} module with two hidden layers, the feature dimensions of which remain unchanged. We utilize the given geographical adjacent matrix.
\end{itemize}

\textbf{Hyper-parameter Setting \& Implementation Details} In our model, we adopt exponential linear unit (ELU)~\cite{clevert2015fast} as nonlinearity for function $g$ in Eq.~\ref{equ:loc-attn-1}, and idendity for function $\phi$ in Eq.~\ref{equ:out}. In the experiment, the input window size is 20 weeks, which spans roughly five months. The hyperparameter $d_{a}$ in the location-aware attention is set to $\frac{D}{2}$ to reduce the number of parameters compared to standard additive attention. The order of the norm $p$ in Eq.~\ref{equ:norm} is set to 2, and $\epsilon$ is 1e-12. The number of filters $K$ is 10 in Eq.~\ref{equ:temporal-conv}.
For all methods using the RNN module, we tune the hidden dimensions of the RNN module from \{10, 20, 30\}, and 20 yields the best performance in most cases. The number of RNN hidden layers and graph layers is optimized to 1 and 2 respectively.
In the training process, the best models are selected by early stopping when the validation accuracy does not increase for 200 consecutive epochs, and the maximum epoch is 1500.
All the parameters are initialized with Glorot initialization \cite{glorot2010understanding} and trained using the Adam \cite{kinga2015method} optimizer with weight decay 5e-4, and dropout rate 0.2. The initial learning rate of all methods is searched from the set \{0.001, 0.005, 0.01\}. The batch size is set to 32 across all datasets. All experimental results are the average of 10 randomized trials. 

Suppose the dimension of weight matrices in graph message passing is set to $D \times D$, the number of parameters of the proposed model is $O(D^2+N^2)$. In our  epidemiological prediction problems, $D$ and $N$ are limited by relatively small numbers.

%% file: sections/result.tex
\section{Results}
\input{tables/short-pcc+rmse.tex}
\input{tables/long-pcc+rmse.tex}

 

 \begin{figure}[h] 
  \centering
  \begin{subfigure}[b]{.9\textwidth}
    \centering
    \includegraphics[width=.9\linewidth]{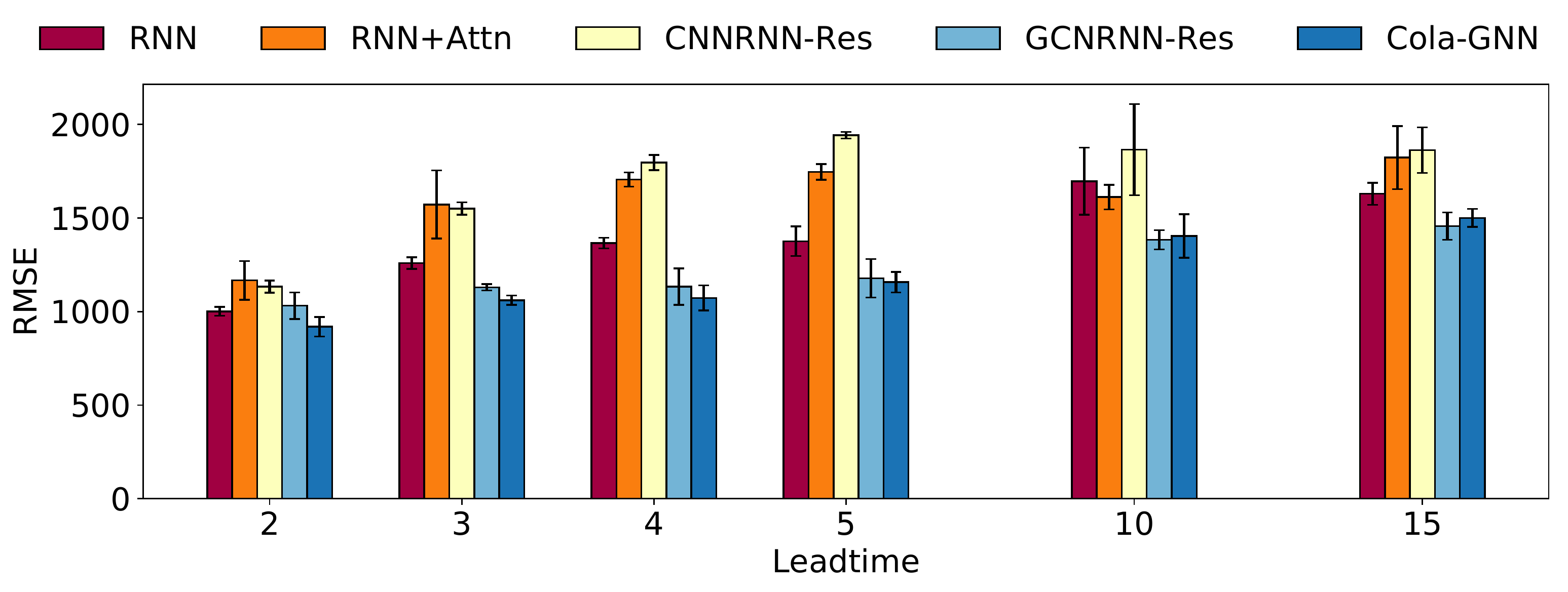}
     \subcaption{Japan-Prefectures}\label{fig:rmse-bar-japan}
  \end{subfigure}
  \begin{subfigure}[b]{.9\textwidth}
    \centering
    \includegraphics[width=.9\linewidth]{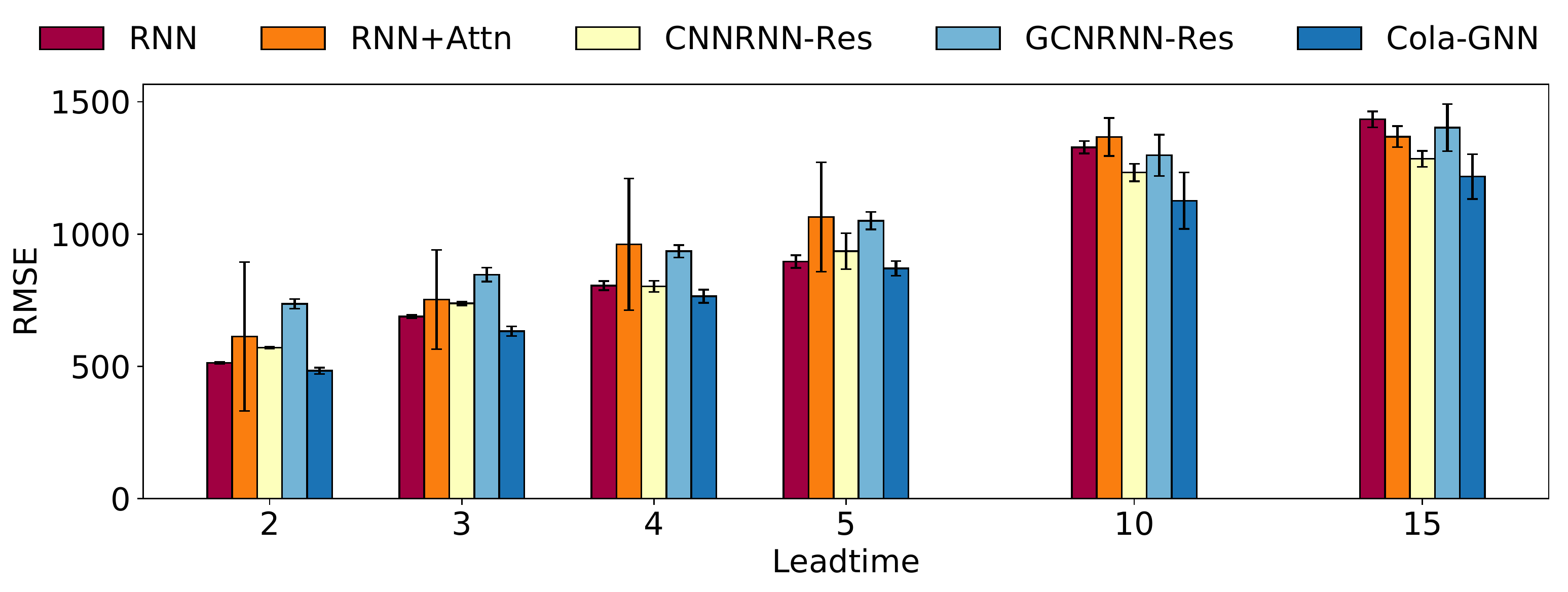}
     \subcaption{US-Regions}\label{fig:rmse-bar-region}
  \end{subfigure}
  \begin{subfigure}[b]{.9\textwidth}
    \centering
    \includegraphics[width=.9\linewidth]{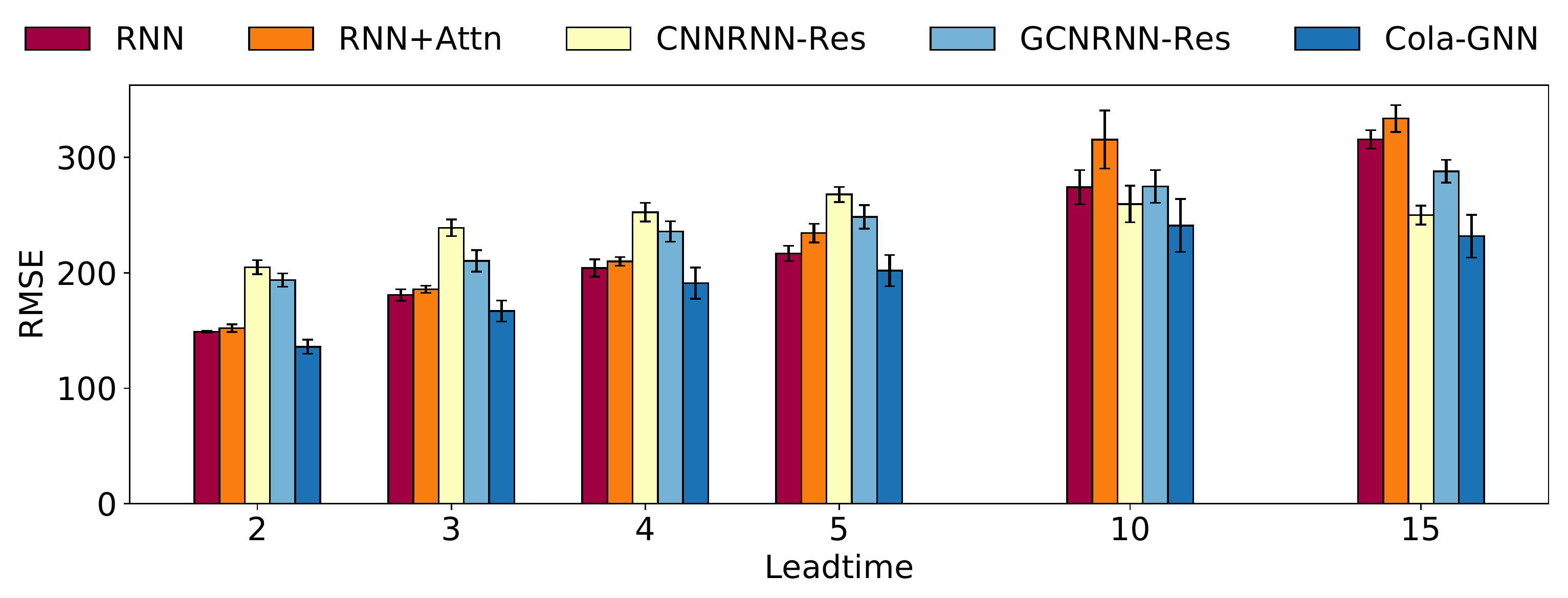}
     \subcaption{US-States}\label{fig:rmse-bar-state}
  \end{subfigure}
  \caption{RMSE of the flu prediction models with different leadtimes on three datasets. } 
  \label{rmse-bar}
 \end{figure}
 
  \begin{figure}[h] 
  \centering
  \begin{subfigure}[b]{.9\textwidth}
    \centering
    \includegraphics[width=.9\linewidth]{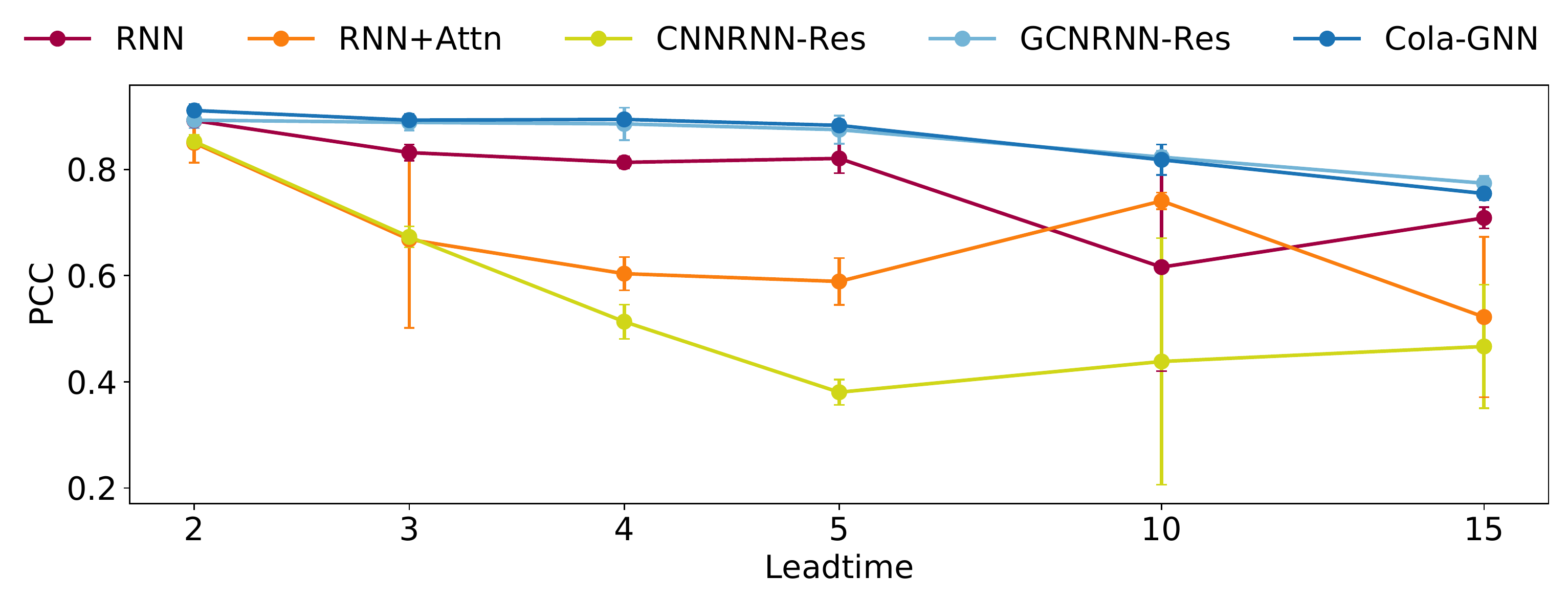}
     \subcaption{Japan-Prefectures}\label{fig:pcc-japan}
  \end{subfigure}
  \begin{subfigure}[b]{.9\textwidth}
    \centering
    \includegraphics[width=.9\linewidth]{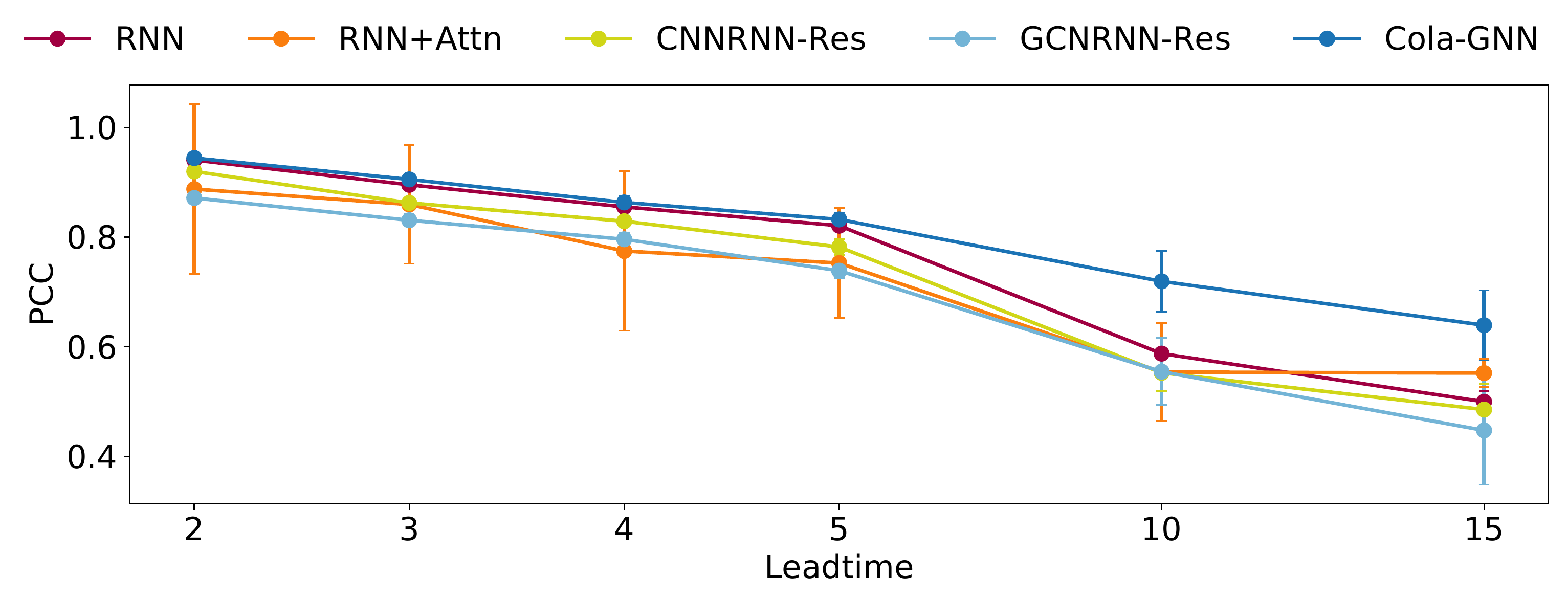}
     \subcaption{US-Regions}\label{fig:pcc-region}
  \end{subfigure}
  \begin{subfigure}[b]{.9\textwidth}
    \centering
    \includegraphics[width=.9\linewidth]{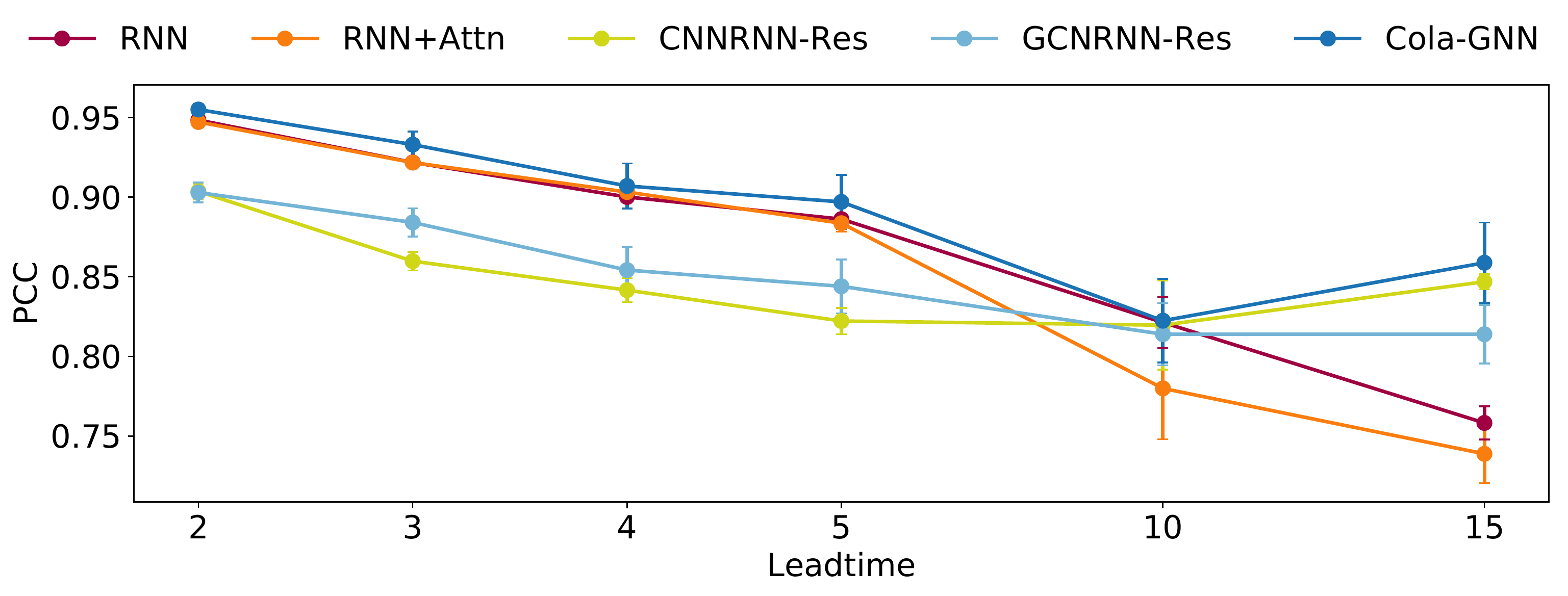}
     \subcaption{US-States}\label{fig:pcc-state}
  \end{subfigure}
  \caption{PCC of the flu prediction models with different leadtimes on three datasets.}
  \label{pcc-curves}
 \end{figure}
 
 
\subsection{Prediction Performance}
We evaluate our approach in short-term (leadtime = 2, 3, 4) and long-term (leadtime = 5, 10, 15) lead time settings. We ignore the case of leadtime = 1, because symptom monitoring data is usually delayed by at least one week. Table~\ref{table:short-rmse-pcc} summarizes the results of all the methods in terms of RMSE and PCC in short-term settings.
We can observe that when the lead time is relatively small, our method achieves the most stable and optimal performance on all datasets. In this case, most of the methods can capture relatively good performance in the three datasets, which is due to the small information gap between the history window and the predicted time, thus the models can fit the temporal pattern more easily. 
The one exception is that in the Japan-Prefectures dataset, the results of most baseline methods deteriorate with a slight increase in lead time. A possible reason for this phenomenon in the Japan-Prefectures dataset is that the seasonal influenza curve in the dataset is less predictive,  even for short-term forecasts. The dataset statistic also shows that Japan-Prefectures dataset has the largest standard deviation.

Table~\ref{table:long-rmse-pcc} reports the RMSE and PCC results in long-term settings. Overall, the proposed method achieves best performance for most datasets with long lead time windows (leadtime = 5, 10 or 15 weeks). Autoregression models have poor performance, especially VAR which has the largest number of model parameters. This suggests the importance of controlling the model complexity for data insufficiency problems. Recurrent neural network models only achieve good predictive performance when lead time is small, which demonstrates that long-term predictions require a better design to capture spatial and temporal dependencies. CNNRNN-Res uses geographic location information and it only performs well in the US-States dataset. In the Japan-Prefectures and US-Regions datasets, the model performs poorly when having long lead time windows. Its variant GCNRNN-Res contains a graph convolutional module that learns the features from adjacent regions. GCNRNN-Res has achieved good results in Japan-Prefectures and US-States datasets. It proves that the graph convolution module can help capture long-term dependencies. The performances of CNNRNN-Res and GCNRNN-Res are unstable on three datasets and often show large variance in multiple rounds of training. To better visualize the results, we show the mean value and standard deviation of the 10 different runs of some models in Figure~\ref{rmse-bar} and Figure~\ref{pcc-curves}. 

If we look at the big picture of the prediction performance, the performance difference of all methods is relatively small when the lead time is 2, but as the lead time increases, the predictive power of simple methods (such as autoregressive) decreases significantly. This suggests that modeling temporal dependence is challenging when a relatively large gap exists between the historical window and the expected prediction time. 

 
\subsection{Case Studies}
\begin{figure}[t]
  \centering 
  \begin{subfigure}[b]{.95\textwidth}
    \centering
    \includegraphics[width=0.95\linewidth]{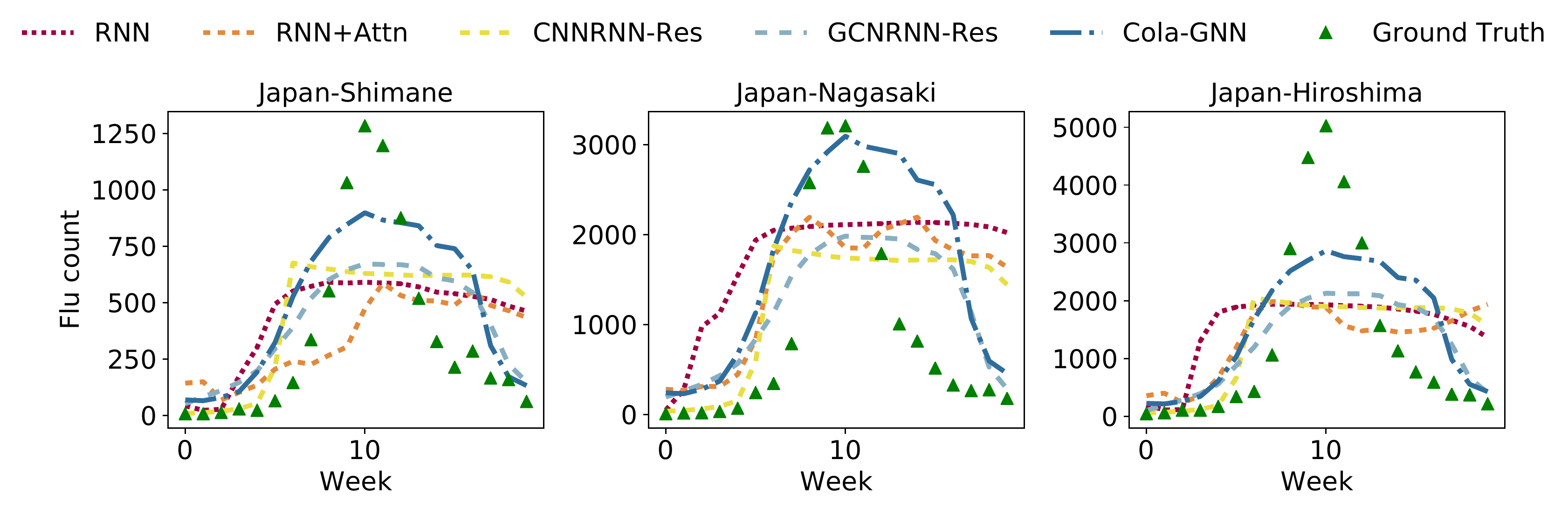}
     \subcaption{Japan-Prefectures}\label{fig:japan-test-curve}
  \end{subfigure} 
    \begin{subfigure}[b]{.95\textwidth}
    \centering
    \includegraphics[width=0.95\linewidth]{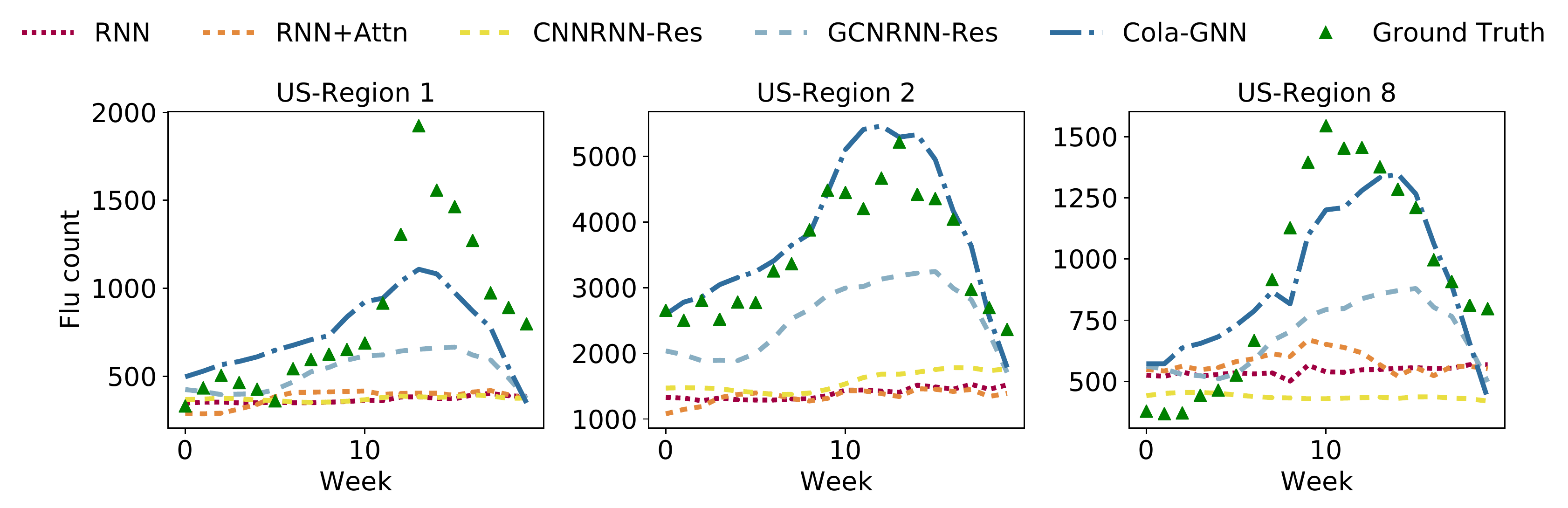}
     \subcaption{US-Regions}\label{fig:region-test-curve}
  \end{subfigure} 
  \begin{subfigure}[b]{.95\textwidth}
    \centering
    \includegraphics[width=0.95\linewidth]{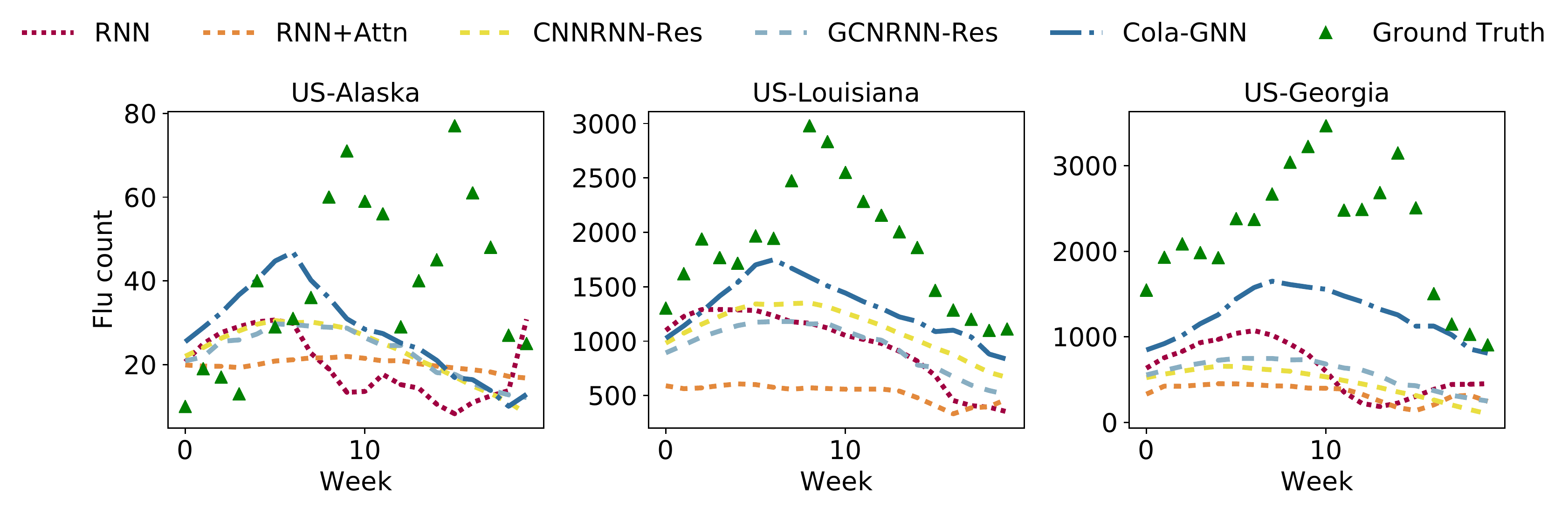}
     \subcaption{US-States}\label{fig:state-test-curve}
  \end{subfigure} 
  \caption{Iterative prediction results when leadtime = 15. We test the models trained in leadtime = 15 by moving the history window. }
  \label{fig:test-curve}
 \end{figure}

To evaluate the long-term predictive performance of the proposed model, we plot a sequence of predictions, where lead time is 15, in the test set. Four better baselines were chosen and the comparison on the three datasets is shown in Figure~\ref{fig:test-curve}. 
We randomly select three locations from each dataset and observe that even though we are using a relatively small window (20) to predict long-term flu count (leadtime = 15), our model is able to better capture the trend and outbreak time of the epidemic outbreak.

    
 
\begin{figure}[t]
  \centering 
  \begin{subfigure}[b]{.95\textwidth}
    \centering
    \includegraphics[width=0.95\linewidth]{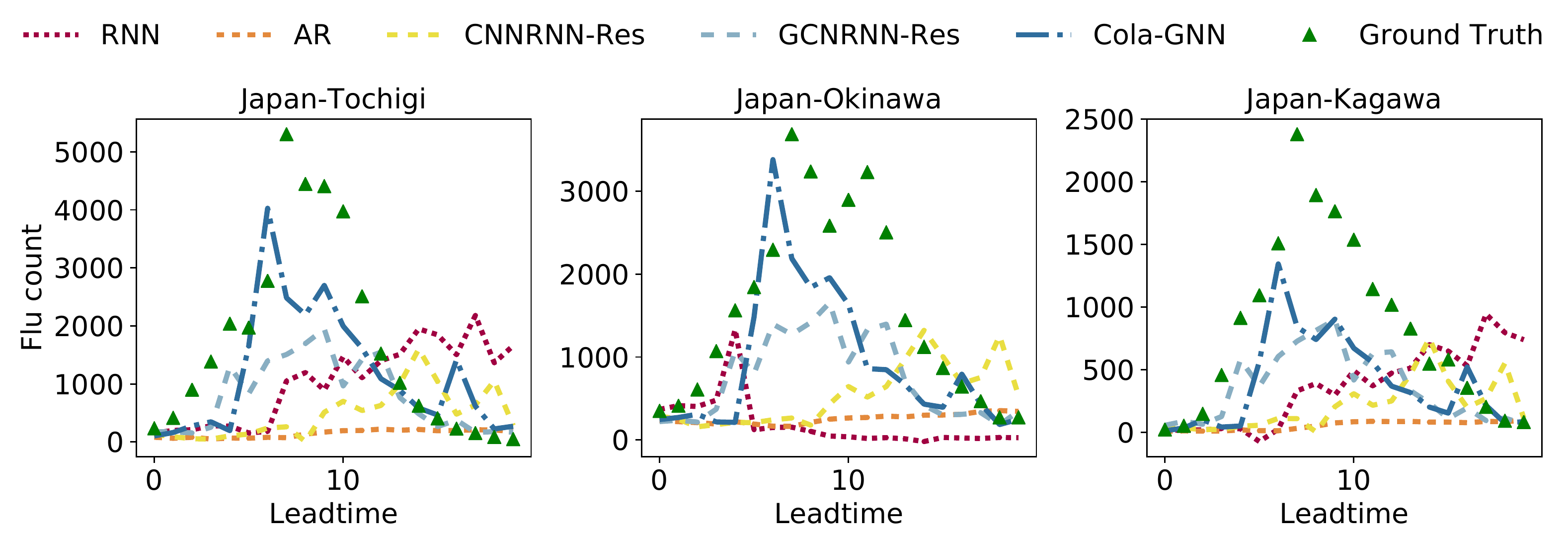}
     \subcaption{Japan-Prefectures}\label{fig:japan-fixw}
  \end{subfigure} 
    \begin{subfigure}[b]{.95\textwidth}
    \centering
    \includegraphics[width=0.95\linewidth]{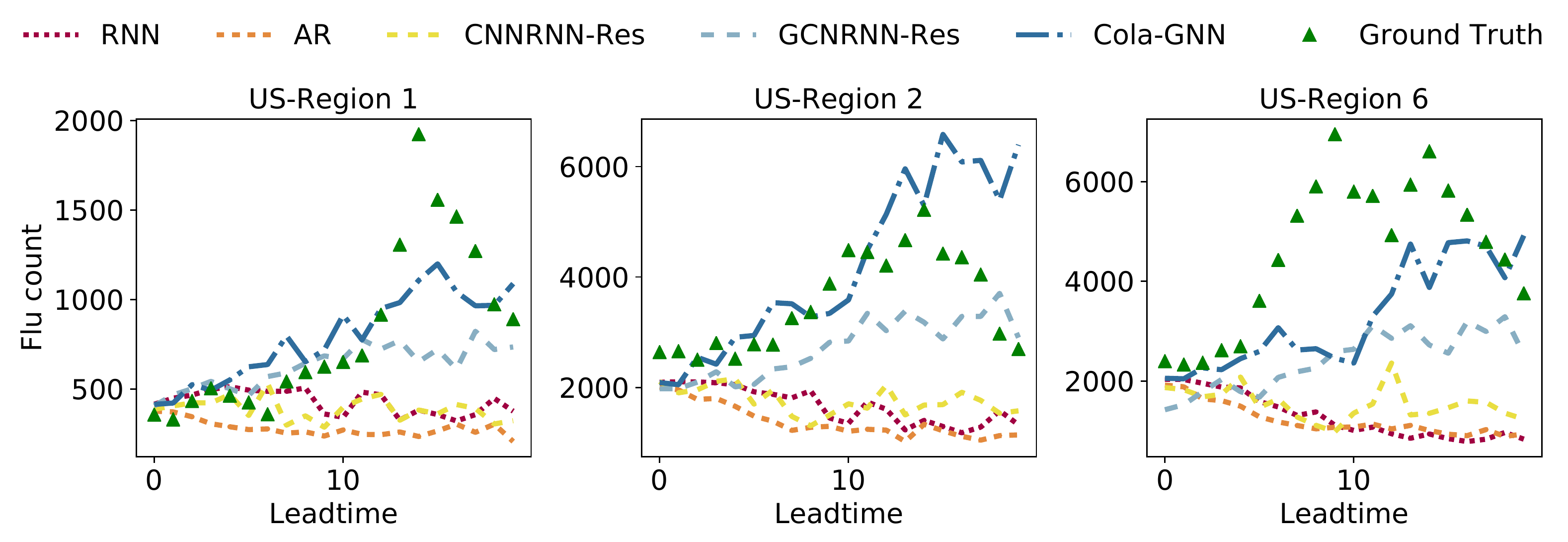}
     \subcaption{US-Regions}\label{fig:region-fixw}
  \end{subfigure} 
  \begin{subfigure}[b]{.95\textwidth}
    \centering
    \includegraphics[width=0.95\linewidth]{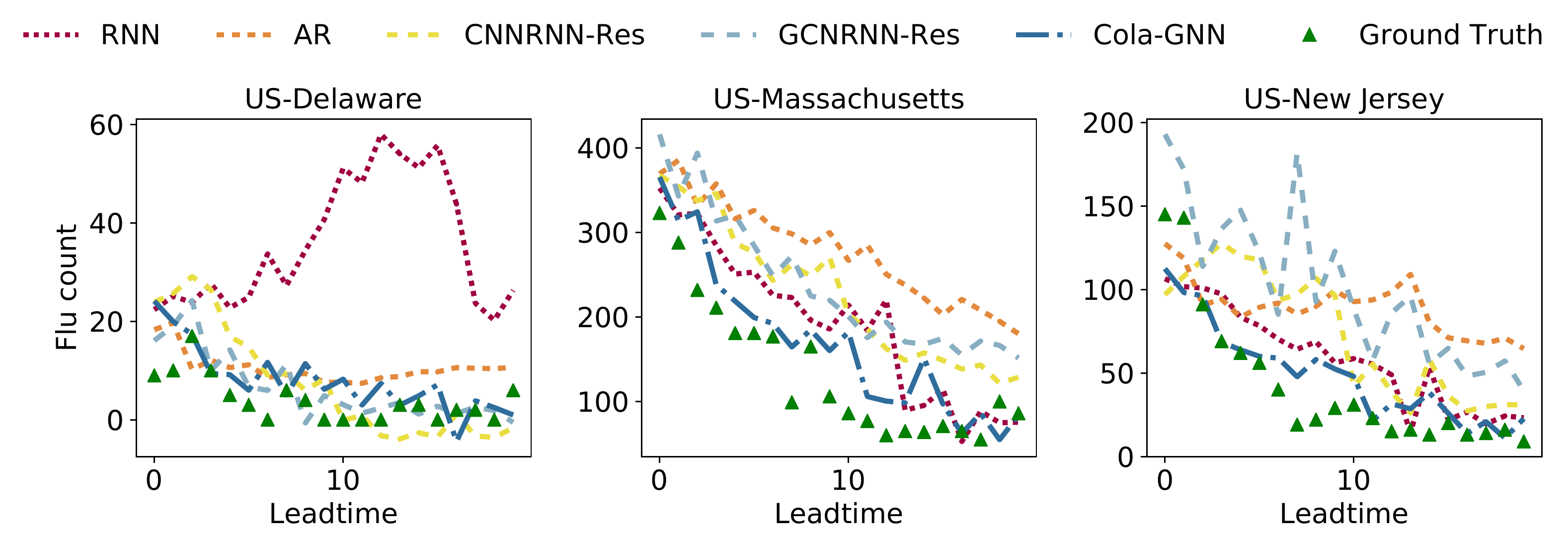}
     \subcaption{US-States}\label{fig:state-fixw}
  \end{subfigure} 
  \caption{Direct prediction curves with fixed input windows. We fix the input window and test the models trained with a lead time from 1 to 20. }
  \label{fig:fix-window-pred}
 \end{figure}

 We fix the input window and plot the prediction curve of leadtime from 1 to 20. 
Likewise, we also randomly sample three locations from each dataset.
 From the observation in Figure~\ref{fig:fix-window-pred}, our model tends to capture the peaks and trends in future time based on given historical data.
 
 
 \subsection{Attention Visualization}
\begin{figure}[t]
  \centering
    \includegraphics[width=.6\linewidth]{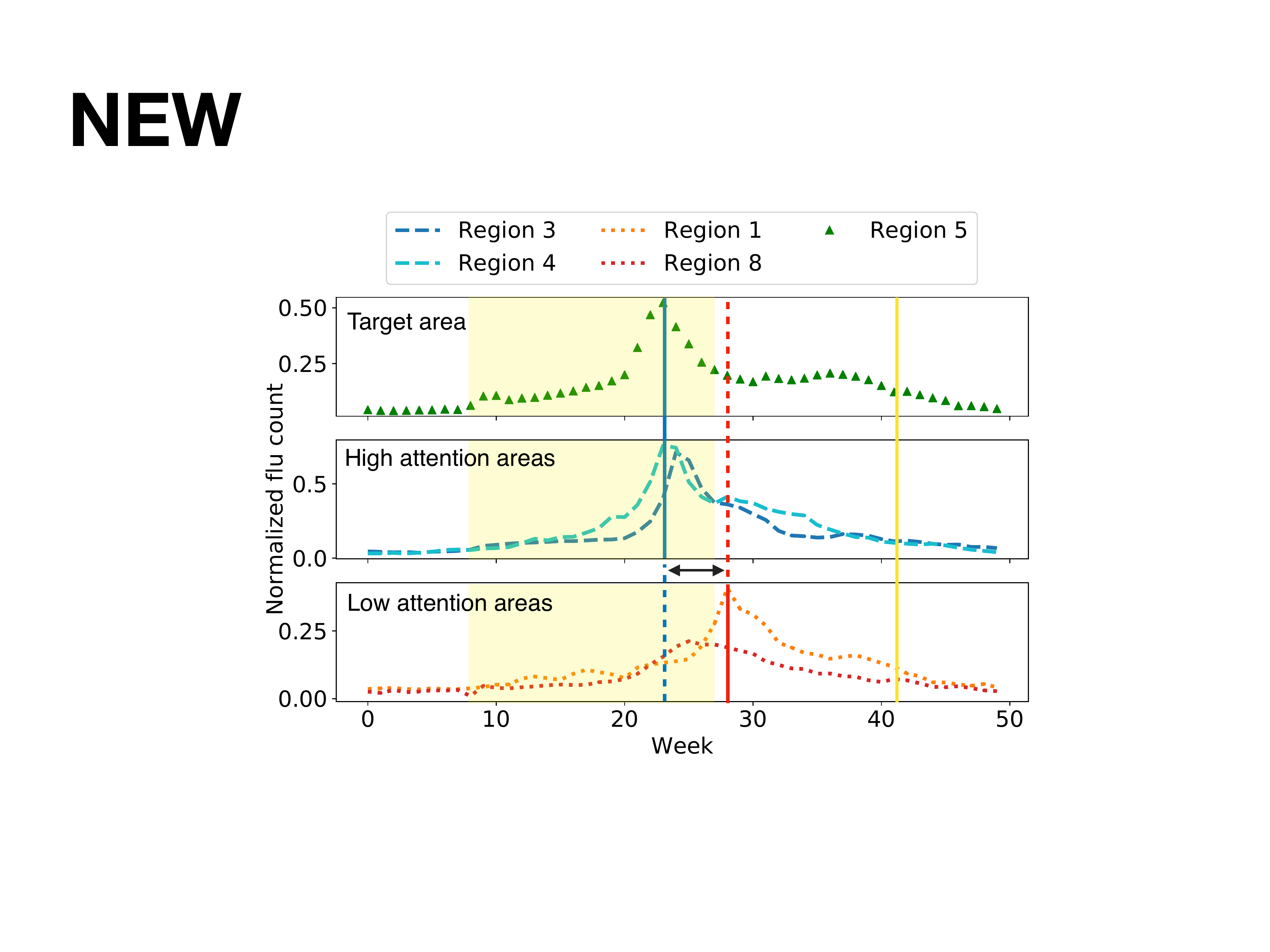}
   \caption{ An example of the location-aware attention mechanism with a lead time of 15 in the US-Regions data set. Yellow line indicates prediction time. Shaded area is the input.}\label{fig:input-sequence}
 \end{figure}

 Figure~\ref{fig:input-sequence} shows an example of the location-aware attention mechanism with a lead time of 15 in the US-Regions dataset. 
In this example, we focus on \textit {region 5}. We visualize the input data of \textit{region 5} and two regions \{\textit{region 3}, \textit{region 4}\} with highest attention values for \textit{region 5}, as well as two regions that has lowest attention values \{\textit{region 1}, \textit{region 8}\}. 
We normalize the data by regions to better compare flu outbreaks across regions.
The time period of the light yellow shade is the input sequence of window = 20. The vertical line indicates the predicted time. We are using only a small part of the sequence of all regions to predict the epidemic outbreak of \textit{region 5} in 15 weeks. The regions with higher attention share same early epidemic outbreak as \textit{region 5} while regions with lower attention values have later outbreak times. 

\begin{figure}[t]
      \centering
    \begin{subfigure}[b]{.3\textwidth}
    \centering
    \includegraphics[width=\linewidth]{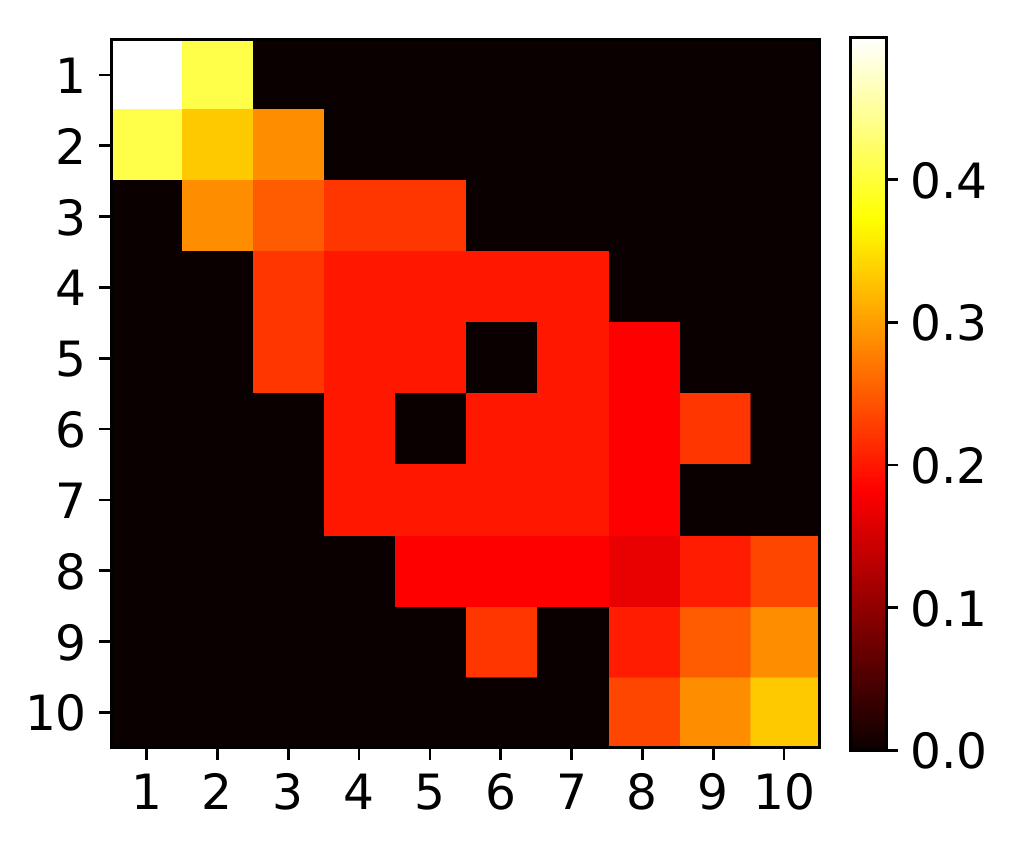}
     \subcaption{Geolocation }\label{fig:geo-mx}
  \end{subfigure} 
  \begin{subfigure}[b]{.3\textwidth}
    \centering
    \includegraphics[width=\linewidth]{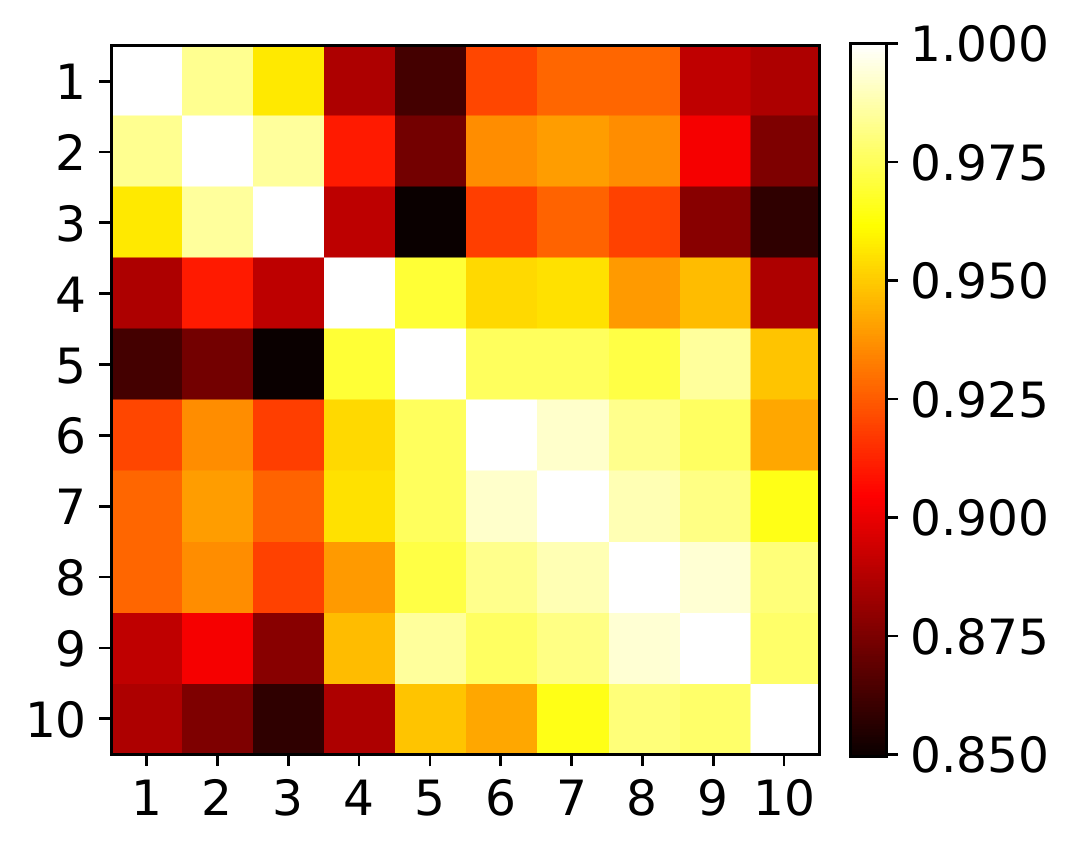}
     \subcaption{Correlation }\label{fig:corr-mx}
  \end{subfigure}
    \begin{subfigure}[b]{.3\textwidth}
    \centering
    \includegraphics[width=\linewidth]{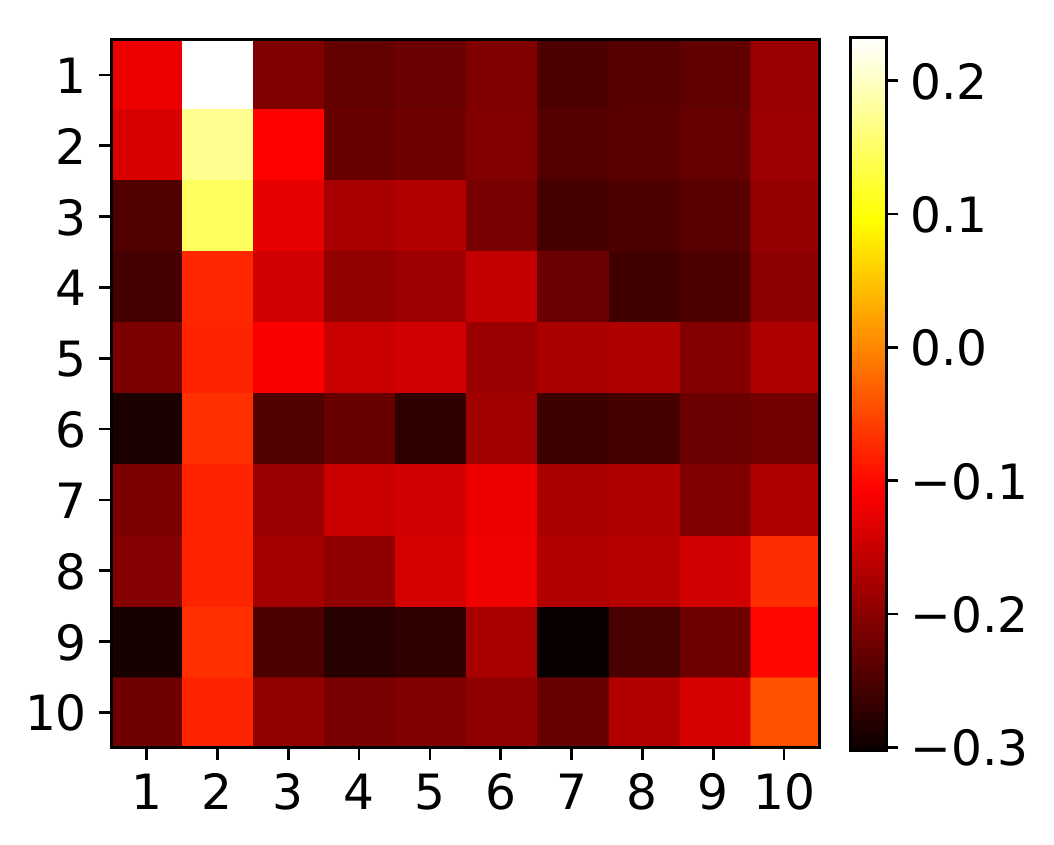}
     \subcaption{Attention }\label{fig:attn-mx}
  \end{subfigure}
  \caption{Comparison of original geolocation matrix (\ref{fig:geo-mx}), input correlation matrix (\ref{fig:corr-mx}), and learned attention matrix (US region).}
  \label{fig:attn-case-study}
 \end{figure}

We show the normalized geolocation distance matrix in Figure~\ref{fig:geo-mx}, which is calculated according to Eq.~\ref {eq:ag-norm}, and the Pearson correlation coefficient of input time series in Figure~\ref{fig:corr-mx}.  The learned attention matrix (Figure~\ref{fig:attn-mx}) utilizes geolocation information as well as additive attentions among regions. From the learned attention matrix, we observe that adjacent regions sometimes get higher attention values. Meanwhile, non-adjacent regions can also receive high attention values given their similar long-term influenza trends. 
The learned attention reveals hidden dynamics (e.g., epidemic outbreaks and peak time) among regions.
 
 \subsection{Ablation Tests}
 \input{tables/new-ablation-pcc+rmse.tex}

 To analyze the effect of each component in our framework, we perform the ablation tests on all the datasets with the follow settings:
\begin{itemize}
\item Cola-GNN w/o $temp$: Remove the temporal convolution module from the proposed model, and use the raw time series input as features in graph message passing.
\item Cola-GNN w/o $loc$: Remove the location-aware attention module and directly use the geographical adjacent matrix which defines the spatial distance between pairs of locations. 

\end{itemize}
The results of RMSE and PCC are shown in Table~\ref{ablation-rmse-pcc}. We can observe that in most cases, variant versions of the proposed method can achieve very good performance. In the US-states dataset, models without temporal or location-aware attention modules are sometimes slightly better than the full model. The US-states dataset has the lowest number of reported influenza cases compare with two other datasets, and the standard deviation is small.
Overall, the full model achieves optimal performance across all datasets.
Note that all datasets are relatively small in size, which means that adding more parameters may affect the performance due to overfitting. However, adding temporal and spatial modules does not change the short-term (leadtime = 2,3,4) prediction very much. Instead, for long-term predictions (leadtime = 15), involving these two modules produces better results. 
  \begin{figure}[t]
  \centering 
    \begin{subfigure}[b]{.45\textwidth}
    \centering
    \includegraphics[width=.9\linewidth]{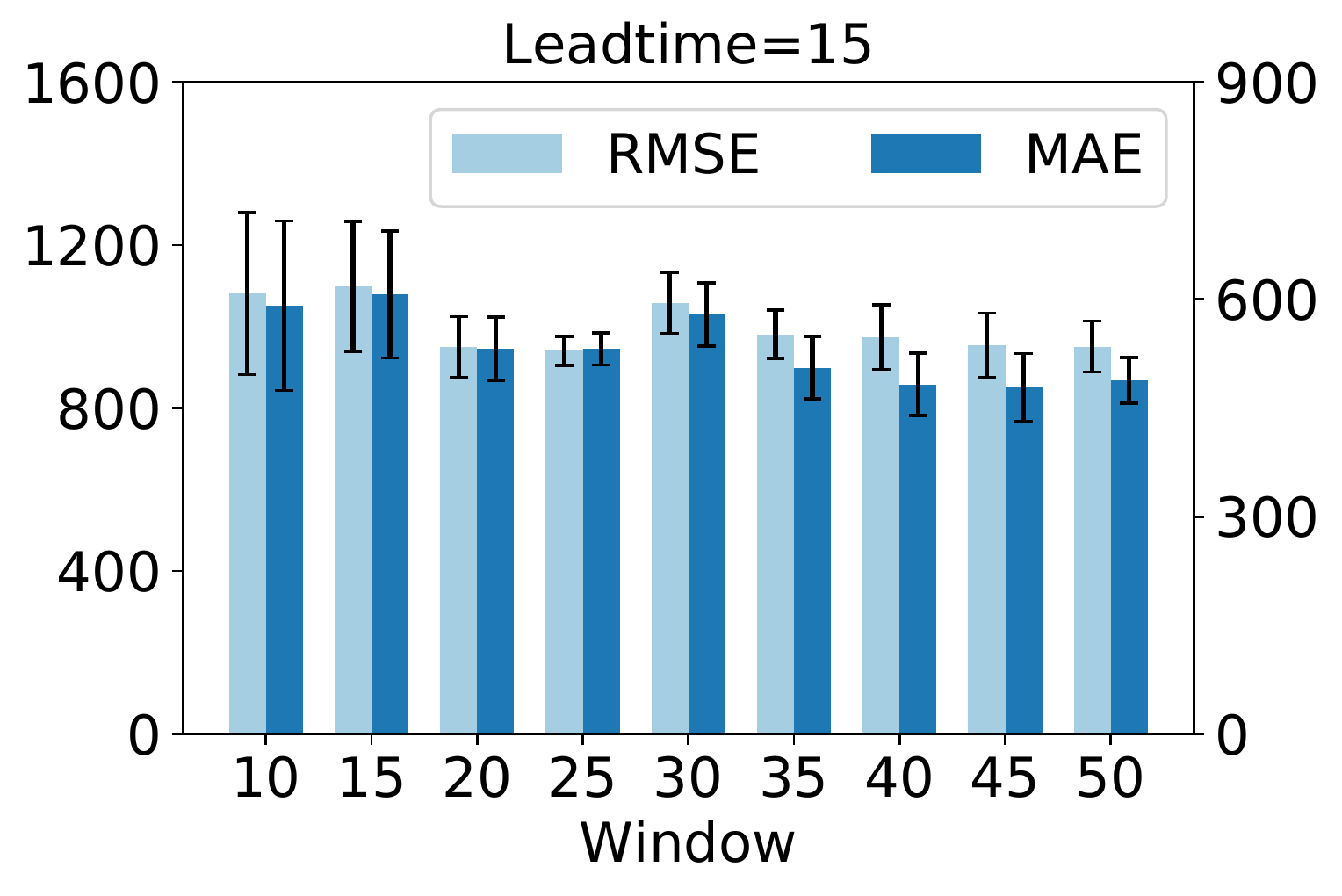}
     \subcaption{US-Regions}\label{fig:region-window}
  \end{subfigure} 
  \hspace{-5pt}
  \begin{subfigure}[b]{.45\textwidth}
    \centering
    \includegraphics[width=.9\linewidth]{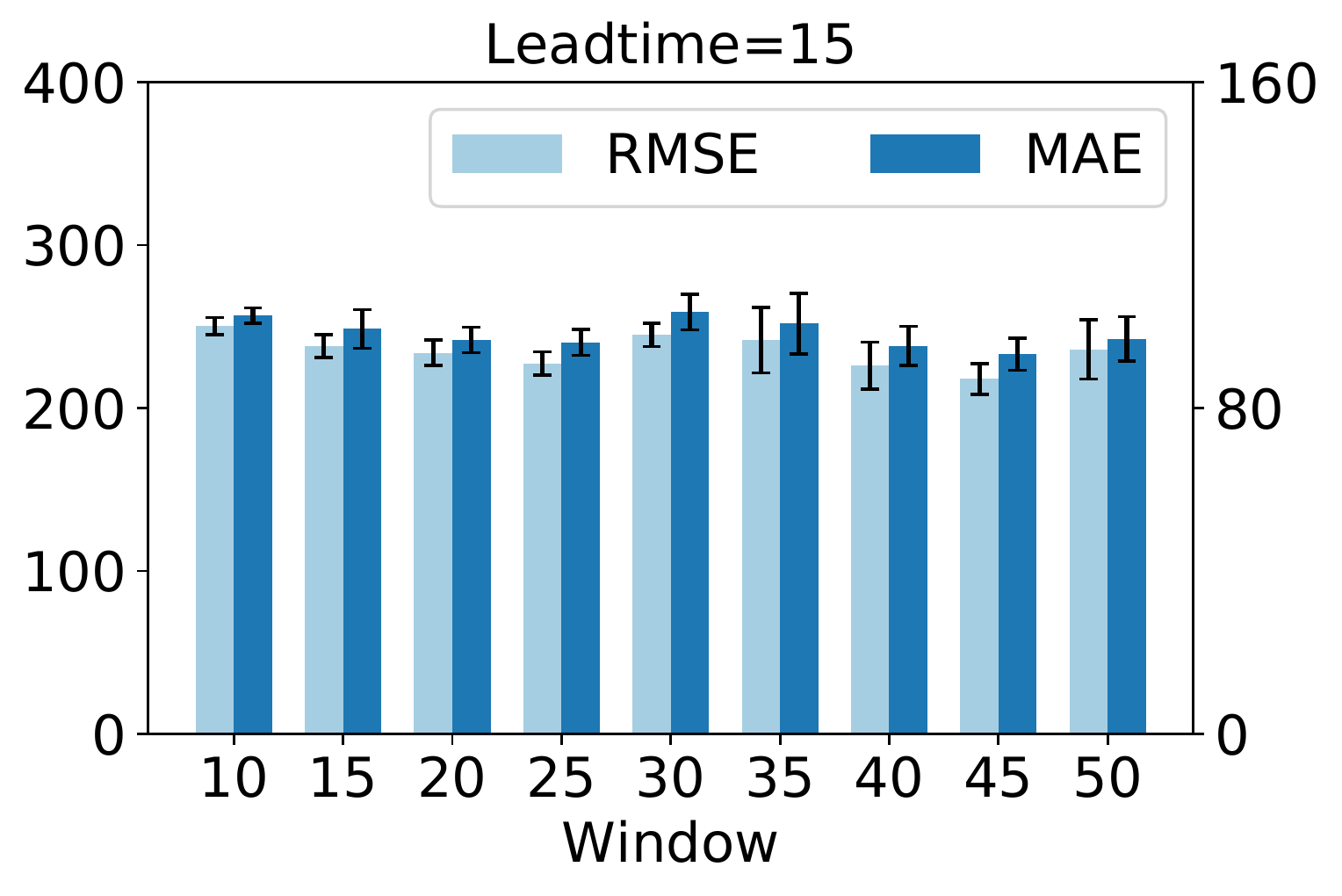}
     \subcaption{US-States}\label{fig:state-window}
  \end{subfigure} 
  \caption{Sensitivity analysis on window size.}
  \label{fig:window-analysis}
 \end{figure}
 
\subsection{Sensitivity Analysis} 
 In this section, we investigate how the prediction performance varies with some hyperparameters. 

  \paragraph{Size of History Windows} To test if our model is sensitive to the length of historical data, we evaluate different window sizes from 10 to 50 with step 5. The experiment was conducted on US-Regions and US-States datasets as shown in Figure~\ref{fig:window-analysis}. 
The predictive performance in RMSE and MAE with different window sizes are fairly stable. We can avoid training with very long sequences and achieve relatively comparable results.

\begin{figure}[t]
  \centering  
    \includegraphics[width=.62\linewidth]{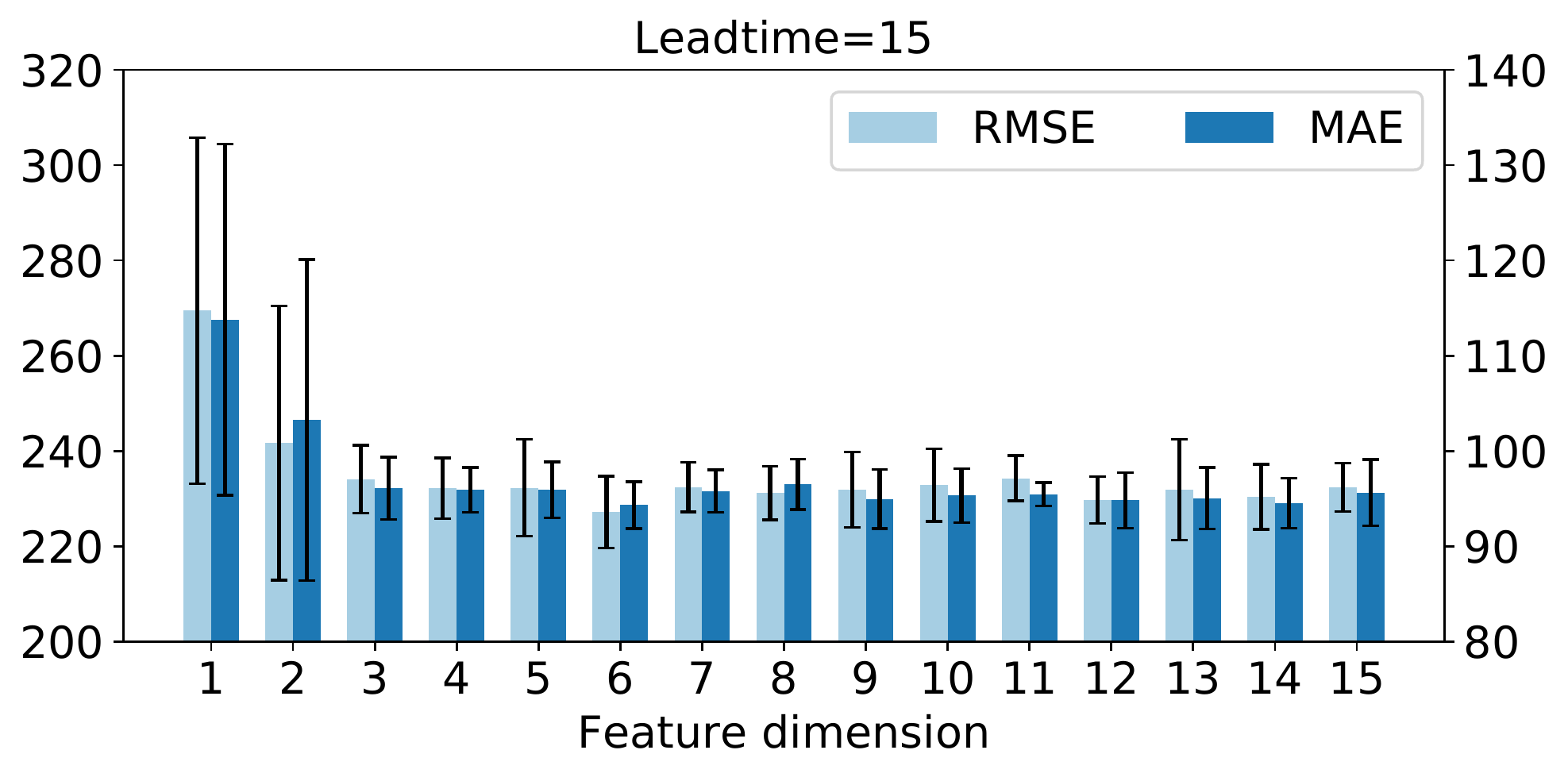}
  \caption{Sensitivity analysis of graph feature size. 
  }
  \label{fig:dim-analysis}
 \end{figure}

  \paragraph{Size of Graph Features} We learn the RNN hidden states from the historical sequence data $h_{i,T}$ and the graph features $h_{i}^{(l)}$ which involves features of other regions by message passing over location-aware attentions. We vary the dimension of the graph feature from 1 to 15 and evaluate the predictive performance in US-States dataset when leadtime is 15. Figure~\ref{fig:dim-analysis} reports RMSE and MAE results. Features of smaller dimensions result in poor predictive performance due to limited encoding power. The model produces better predictive power when the feature dimension is larger.
  
\begin{figure}[t]
  \centering 
    \begin{subfigure}[b]{.4\textwidth}
    \centering
    \includegraphics[width=.9\linewidth]{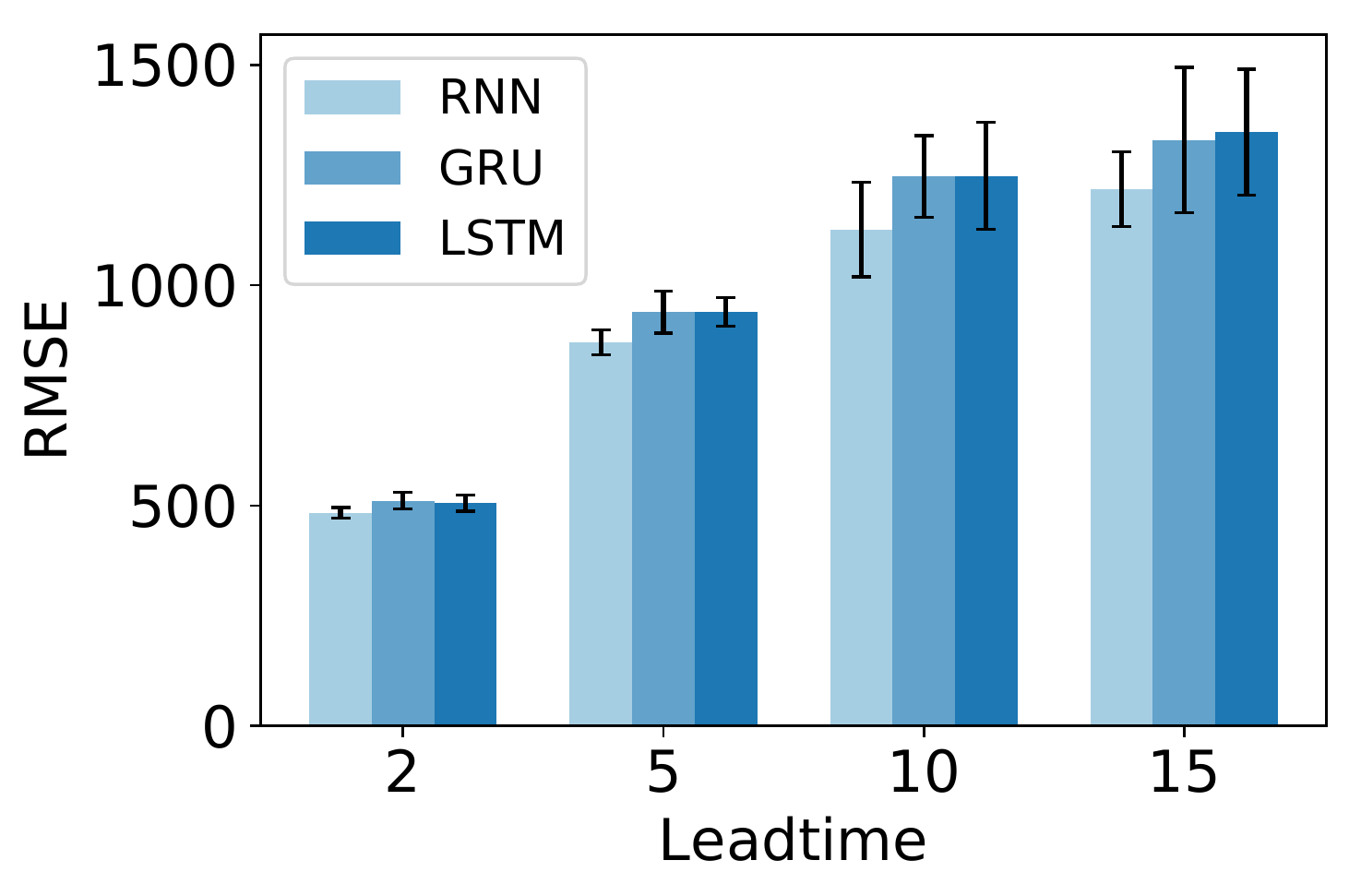}
     \subcaption{US-Regions}
     \label{fig:region-rnn}
  \end{subfigure} 
  \hspace{-5pt}
  \begin{subfigure}[b]{.4\textwidth}
    \centering
    \includegraphics[width=.9\linewidth]{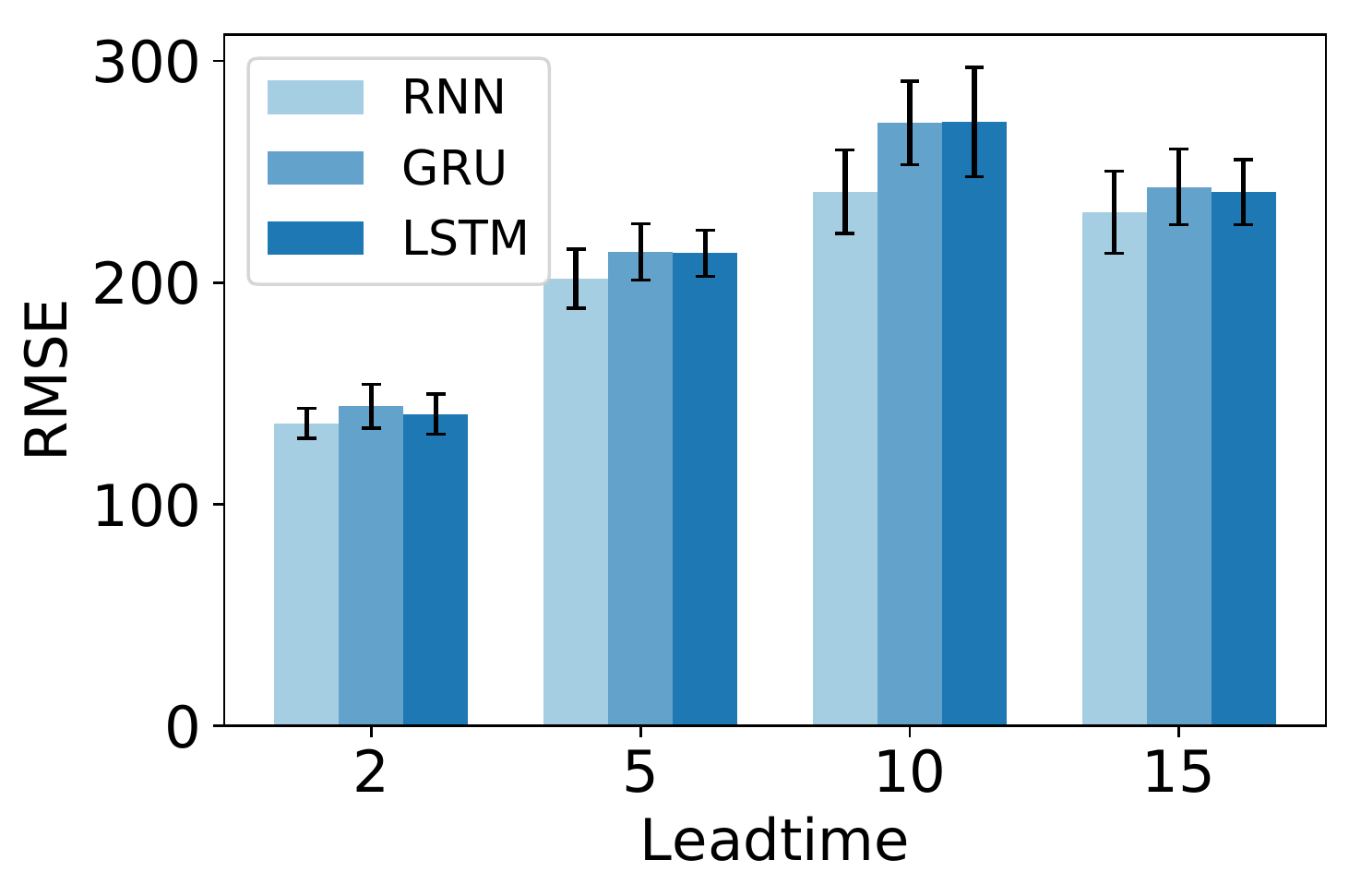}
     \subcaption{US-States}
     \label{fig:state-rnn}
  \end{subfigure} 
  \caption{Sensitivity analysis on RNN modules.}
  \label{fig:rnn-analysis}
 \end{figure}

  \paragraph{RNN Modules} 
The RNN module is used to output a hidden state vector for each location based on given historical data. The hidden state vector is then provided to the location-aware attention module. We replaced the RNN modules with GRU and LSTM to assess their impact on model performance. Figure~\ref{fig:rnn-analysis} shows RMSE results for leadtime = 2,5,10,15 in US-Regions and US-States datasets. We found that the performance of GRU and LSTM is not better than a simple RNN. The likely reason is that they involve more model parameters and tend to overfit in the epidemiological datasets.
  
\subsection{Model Complexity} 
\begin{table}[h]
\footnotesize
\centering
\caption{Runtime comparison of models on the  US-States dataset. Runtime is the time spent on a single GPU per epoch.}\label{complexity}
\begin{tabular}{lrc}
\toprule
\textbf{Architecture}    & \textbf{Parameters}  & \textbf{Runtime(s)}  \\
\midrule
GAR      &     21        &  0.01           \\ 
AR      &      1,029    &   0.02 \\ 
VAR      &     48,069        &  0.02  \\ 
ARMA      &      1,960 & 0.03          \\ 
RNN      &     481       &   0.04   \\ 
RNN+Attn   &  1,321    &    0.58 \\
CNNRNN-Res   &    7,695    &  0.04     \\
GCNRNN-Res    &    6,214     &  0.04      \\
Cola-GNN  &    3,778    &    0.21     \\
\bottomrule
\end{tabular}
\end{table}
Table~\ref{complexity} shows the comparison of runtimes and numbers of parameters for each model on the US-States dataset, which has the largest number of \textit{regions} among the three datasets. In this task, all methods can be effectively trained due to the nature of the datasets. Meanwhile, we only utilize flu disease data and geographic location data, while ignoring other external features. Compared with other methods, the proposed method has no significant effect on training efficiency. It can also control the size of the model parameters to prevent overfitting.

%% file: tables/short-pcc+rmse.tex
\begin{table*}[t]
\footnotesize
\centering
\caption{RMSE and PCC performance of different methods on the three datasets with leadtime = 2, 3, 4. Bold face indicates the best result of each column and underlined the second-best. \textbf{(Short-term)}}\label{table:short-rmse-pcc}
\scalebox{0.95}{\begin{tabular}{lccccccccc}
\toprule
 & \multicolumn{3}{c}{\textbf{Japan-Prefectures}} & \multicolumn{3}{c}{\textbf{US-Regions}} & \multicolumn{3}{c}{\textbf{US-States}} \\
\cmidrule(r){2-4}\cmidrule(r){5-7}\cmidrule(r){8-10}
\textbf{RMSE($\downarrow$)} & 2 & 3 & 4 & 2 & 3 & 4 & 2 & 3 & 4  \\
\hline
  
GAR         & 1232         & 1628          & 1865          & 536          & 715          & 859          & 150          & 187          & 213          \\
AR          & 1377         & 1705          & 1901          & 570          & 757          & 888          & 161          & 204          & 231          \\
VAR         & 1361         & 1711          & 1910          & 741          & 870          & 967          & 290          & 276          & 283          \\
ARMA        & 1371         & 1703          & 1902          & 560          & 742          & 874          & 161          & 200          & 228          \\
RNN         & 1001         & 1259          & 1366          &\underline{513}    &\underline{689}    & 805          &\underline{149}    &\underline{181}    &\underline{204}    \\
RNN+Attn    & 1166         & 1572          & 1706          & 613          & 753          & 962          & 152          & 186          & 210          \\
CNNRNN-Res & 1133         & 1550          & 1795          & 571          & 738          &\underline{802}    & 205          & 239          & 253          \\
GCNRNN-Res &\underline{1031}   &\underline{1129}    &\underline{1133}    & 736          & 847          & 935          & 194          & 210          & 236          \\
Cola-GNN    & \textbf{919} & \textbf{1060} & \textbf{1072} & \textbf{483} & \textbf{633} & \textbf{765} & \textbf{136} & \textbf{167} & \textbf{191} \\

\midrule
\textbf{PCC($\uparrow$)}  & 2 & 3 & 4 & 2 & 3 & 4 & 2 & 3 & 4  \\
\midrule
GAR         & 0.804          & 0.626          & 0.461          & 0.932          & 0.881          & 0.835          & 0.945         & 0.914          & 0.893         \\
AR          & 0.752           & 0.579          & 0.428          & 0.927          & 0.878          & 0.834          & 0.94         & 0.909          & 0.885          \\
VAR         & 0.754          & 0.585          & 0.419          & 0.859          & 0.797          & 0.741          & 0.765         & 0.79          & 0.78         \\
ARMA        & 0.754           & 0.579          & 0.428          & 0.927          & 0.876          & 0.833          & 0.939         & 0.909          & 0.886         \\
RNN         & 0.892           & 0.833          & 0.813          &\underline{0.94}    &\underline{ 0.895}    &\underline{0.855}    &\underline{ 0.948}   &\underline{0.922}    & 0.9         \\
RNN+Attn    & 0.85          & 0.668          & 0.604          & 0.887          & 0.859          & 0.774          & 0.947         & 0.922          &\underline{ 0.903}   \\
CNNRNN-Res & 0.852          & 0.673          & 0.513          & 0.92          & 0.862          & 0.829          & 0.904         & 0.86          & 0.842         \\
GCNRNN-Res &\underline{0.893}    &\underline{ 0.889}    &\underline{0.886}    & 0.871          & 0.831          & 0.796          & 0.903         & 0.884          & 0.854         \\
Cola-GNN    & \textbf{0.911} & \textbf{0.893} & \textbf{0.894} & \textbf{0.944} & \textbf{0.905} & \textbf{0.863} & \textbf{0.955} & \textbf{0.933} & \textbf{0.907} \\

\bottomrule
\end{tabular}}
\end{table*}

%% file: tables/long-pcc+rmse.tex
\begin{table*}[t]
\footnotesize
\centering
\caption{RMSE and PCC performance of different methods on the three datasets with leadtime = 5, 10, 15. Bold face indicates the best result of each column and underlined the second-best. \textbf{(Long-term)}}\label{table:long-rmse-pcc}
\scalebox{0.95}{\begin{tabular}{lccccccccc}
\toprule
 & \multicolumn{3}{c}{\textbf{Japan-Prefectures}} & \multicolumn{3}{c}{\textbf{US-Regions}} & \multicolumn{3}{c}{\textbf{US-States}} \\
\cmidrule(r){2-4}\cmidrule(r){5-7}\cmidrule(r){8-10}
\textbf{RMSE($\downarrow$)} & 5 & 10 & 15 & 5 & 10 & 15 & 5 & 10 & 15  \\
\hline
  
GAR         & 1988          & 2065          & 2016          & 991          & 1377          & 1465          & 236          & 314          & 340          \\
AR          & 2013          & 2107          & 2042          & 997          & 1330          & 1404          & 251          & 306          & 327          \\
VAR         & 2025          & 1942          & 1899          & 1059         & 1270          & 1299          & 295          & 324          & 352          \\
ARMA        & 2013          & 2105          & 2041          & 989          & 1322          & 1400          & 250          & 306          & 326          \\
RNN         & 1376          & 1696          & 1629          &\underline{ 896}    & 1328          & 1434          &\underline{ 217}    & 274          & 315          \\
RNN+Attn    & 1746          & 1612          & 1823          & 1065         & 1367          & 1368          & 234          & 315          & 334          \\
CNNRNN-Res & 1942          & 1865          & 1862          & 936          &\underline{1233}    & \underline{1285}    & 267          & \underline{260}    & \underline{250}    \\
GCNRNN-Res &\underline{ 1178}    & \textbf{1384} & \textbf{1457} & 1051         & 1298          & 1402          & 248          & 275          & 288          \\
Cola-GNN    & \textbf{1156} & \underline{1403}    & \underline{ 1500}    & \textbf{871} & \textbf{1126} & \textbf{1218} & \textbf{202} & \textbf{241} & \textbf{232} \\

\midrule
\textbf{PCC($\uparrow$)}  & 5 & 10 & 15 & 5 & 10 & 15 & 5 & 10 & 15  \\
\midrule

GAR         & 0.339          & 0.288          & 0.47         & 0.79          & 0.581          & 0.485         & 0.875          & 0.777          & 0.742          \\
AR          & 0.310         & 0.238          & 0.483         & 0.792          & 0.612          & \underline{0.527}   & 0.863          & 0.773          & 0.723          \\
VAR         & 0.3         & 0.426          & 0.474         & 0.685          & 0.508          & 0.467         & 0.758          & 0.709          & 0.6529          \\
ARMA        & 0.31          & 0.253          & 0.486         & 0.792          & \underline{0.614}    & 0.52         & 0.862          & 0.773           & 0.725          \\
RNN         & 0.821          & 0.616          & 0.709         & \underline{0.821}    & 0.587          & 0.499         & \textbf{0.886} & \underline{0.821}    & 0.758          \\
RNN+Attn    & 0.59           & 0.741          & 0.522          & 0.752          & 0.554          & 0.552         & \underline{0.884}    & 0.78          & 0.739          \\
CNNRNN-Res & 0.38          & 0.438          & 0.467         & 0.782          & 0.552          & 0.4851         & 0.822          & 0.82          &\underline{0.847}    \\
GCNRNN-Res & \underline{0.875}    & \textbf{0.823} & \textbf{0.774} & 0.739          & 0.554          & 0.4471         & 0.844           & 0.814          & 0.814          \\
Cola-GNN    & \textbf{0.883} &\underline{0.818}    & \underline{0.754}   & \textbf{0.832} & \textbf{0.719} & \textbf{0.639} & 0.897          & \textbf{0.822} & \textbf{0.859} \\

\bottomrule
\end{tabular}}
\end{table*}

%% file: tables/new-ablation-pcc+rmse.tex
\begin{table}[t]
\centering
\caption{Ablation test results in RMSE(top) and PCC(bottom) when leadtime=2,3,4,5,10,15 for three datasets.}\label{ablation-rmse-pcc}
\scalebox{1.0}{\begin{tabular}{lcccccc}
\toprule
\textbf{RMSE($\downarrow$)} & 2 & 3 & 4 & 5 & 10 & 15 \\
\midrule
& \multicolumn{6}{c}{Japan-Prefectures} \\
\cmidrule(r){2-7}
Cola-GNN w/o $temp$  & 911  & 1115& 1204 & 1310 & 1388& 1517  \\ 
Cola-GNN w/o $loc$ & 942 &1154 & 1164& 1195 & 1473& 1576   \\ 
Cola-GNN  &919 &1060& 1072 &1156 &1403   & 1500    \\ 
\midrule
& \multicolumn{6}{c}{US-Regions}  \\
\cmidrule(r){2-7}
Cola-GNN w/o $temp$  & 485 & 662&772 & 888 &1144 & 1228   \\ 
Cola-GNN w/o $loc$  & 499  & 666&782 & 890 & 1179& 1292   \\  
Cola-GNN & 483& 633 & 765 &871 & 1126 &1218   \\ 
\midrule
& \multicolumn{6}{c}{US-States} \\
\cmidrule(r){2-7}
Cola-GNN w/o $temp$ & 138  & 169 & 188 & 194 & 251 &251   \\  
Cola-GNN w/o $loc$ & 138 &169 &193 &202  &246 &246   \\   
Cola-GNN & 136 & 167 & 191 &202 &241 &232   \\ 
\midrule 
\textbf{PCC($\uparrow$)}  & 2 & 3 & 4 & 5 & 10 & 15 \\
\midrule 
& \multicolumn{6}{c}{Japan-Prefectures} \\
\cmidrule(r){2-7}
Cola-GNN w/o $temp$ &0.91  & 0.867 &0.846 & 0.818 & 0.793& 0.744   \\ 
Cola-GNN w/o $loc$ &0.914 & 0.881& 0.89& 0.88& 0.781& 0.727   \\ 
Cola-GNN &  0.911 & 0.893 & 0.894 & 0.883 &0.818   & 0.754  \\ 
\midrule
& \multicolumn{6}{c}{US-Regions} \\
\cmidrule(r){2-7}
Cola-GNN w/o $temp$ & 0.944 & 0.902& 0.861& 0.824& 0.712& 0.588   \\ 
Cola-GNN w/o $loc$& 0.942& 0.898 & 0.858& 0.824& 0.682& 0.582    \\ 
Cola-GNN  &0.944 &0.905 &0.863 &0.832 & 0.719 & 0.639    \\ 
\midrule
& \multicolumn{6}{c}{US-States} \\
\cmidrule(r){2-7}
Cola-GNN w/o $temp$ & 0.953 & 0.93 & 0.908 & 0.908 & 0.833 &  0.836     \\ 
Cola-GNN w/o $loc$ & 0.955 & 0.931 & 0.913 &0.904 &0.856 & 0.855  \\ 
Cola-GNN & 0.955 & 0.933 & 0.907 & 0.897  & 0.822 & 0.859    \\ 
\bottomrule
\end{tabular}}
\label{tbl:ablation-rmse}
\end{table}

%% file: main.bbl
\begin{thebibliography}{32}
\providecommand{\natexlab}[1]{#1}
\providecommand{\url}[1]{\texttt{#1}}
\expandafter\ifx\csname urlstyle\endcsname\relax
  \providecommand{\doi}[1]{doi: #1}\else
  \providecommand{\doi}{doi: \begingroup \urlstyle{rm}\Url}\fi

\bibitem[Achrekar et~al.(2011)Achrekar, Gandhe, Lazarus, Yu, and
  Liu]{achrekar2011predicting}
Harshavardhan Achrekar, Avinash Gandhe, Ross Lazarus, Ssu-Hsin Yu, and Benyuan
  Liu.
\newblock Predicting flu trends using twitter data.
\newblock In \emph{2011 IEEE conference on computer communications workshops
  (INFOCOM WKSHPS)}, pages 702--707. IEEE, 2011.

\bibitem[Bahdanau et~al.(2014)Bahdanau, Cho, and Bengio]{bahdanau2014neural}
Dzmitry Bahdanau, Kyunghyun Cho, and Yoshua Bengio.
\newblock Neural machine translation by jointly learning to align and
  translate.
\newblock \emph{arXiv preprint arXiv:1409.0473}, 2014.

\bibitem[Bisset et~al.(2009)Bisset, Chen, Feng, Kumar, and
  Marathe]{bisset2009epifast}
Keith~R Bisset, Jiangzhuo Chen, Xizhou Feng, VS~Kumar, and Madhav~V Marathe.
\newblock Epifast: a fast algorithm for large scale realistic epidemic
  simulations on distributed memory systems.
\newblock In \emph{Proceedings of the 23rd international conference on
  Supercomputing}, pages 430--439. ACM, 2009.

\bibitem[Brownstein et~al.(2017)Brownstein, Chu, Marathe, Marathe, Nguyen,
  Paolotti, Perra, Perrotta, Santillana, Swarup,
  et~al.]{brownstein2017combining}
John~S Brownstein, Shuyu Chu, Achla Marathe, Madhav~V Marathe, Andre~T Nguyen,
  Daniela Paolotti, Nicola Perra, Daniela Perrotta, Mauricio Santillana,
  Samarth Swarup, et~al.
\newblock Combining participatory influenza surveillance with modeling and
  forecasting: Three alternative approaches.
\newblock \emph{JMIR public health and surveillance}, 3\penalty0 (4):\penalty0
  e83, 2017.

\bibitem[Chakraborty et~al.(2014)Chakraborty, Khadivi, Lewis, Mahendiran, Chen,
  Butler, Nsoesie, Mekaru, Brownstein, Marathe,
  et~al.]{chakraborty2014forecasting}
Prithwish Chakraborty, Pejman Khadivi, Bryan Lewis, Aravindan Mahendiran,
  Jiangzhuo Chen, Patrick Butler, Elaine~O Nsoesie, Sumiko~R Mekaru, John~S
  Brownstein, Madhav~V Marathe, et~al.
\newblock Forecasting a moving target: Ensemble models for ili case count
  predictions.
\newblock In \emph{Proceedings of the 2014 SIAM international conference on
  data mining}, pages 262--270. SIAM, 2014.

\bibitem[Cheng et~al.(2016)Cheng, Dong, and Lapata]{cheng2016long}
Jianpeng Cheng, Li~Dong, and Mirella Lapata.
\newblock Long short-term memory-networks for machine reading.
\newblock In \emph{Proceedings of the 2016 Conference on Empirical Methods in
  Natural Language Processing}, pages 551--561, Austin, Texas, November 2016.
  ACL.
\newblock \doi{10.18653/v1/D16-1053}.

\bibitem[Cho et~al.(2014)Cho, van Merri{\"e}nboer, Gulcehre, Bahdanau,
  Bougares, Schwenk, and Bengio]{cho2014learning}
Kyunghyun Cho, Bart van Merri{\"e}nboer, Caglar Gulcehre, Dzmitry Bahdanau,
  Fethi Bougares, Holger Schwenk, and Yoshua Bengio.
\newblock Learning phrase representations using {RNN} encoder{--}decoder for
  statistical machine translation.
\newblock In \emph{Proceedings of the 2014 Conference on Empirical Methods in
  Natural Language Processing ({EMNLP})}, pages 1724--1734, Doha, Qatar,
  October 2014. ACL.
\newblock \doi{10.3115/v1/D14-1179}.

\bibitem[Chowell et~al.(2008)Chowell, Miller, and Viboud]{chowell2008seasonal}
GMAM Chowell, MA~Miller, and C~Viboud.
\newblock Seasonal influenza in the united states, france, and australia:
  transmission and prospects for control.
\newblock \emph{Epidemiology \& Infection}, 136\penalty0 (6):\penalty0
  852--864, 2008.

\bibitem[Clevert et~al.(2015)Clevert, Unterthiner, and
  Hochreiter]{clevert2015fast}
Djork-Arn{\'e} Clevert, Thomas Unterthiner, and Sepp Hochreiter.
\newblock Fast and accurate deep network learning by exponential linear units
  (elus).
\newblock In \emph{Proceedings of the 2015 International Conference on Learning
  Representations}, volume abs/1511.07289, 2015.

\bibitem[Du et~al.(2014)Du, Xu, Song, Ding, and Chu]{du2014prediction}
Baoxiang Du, Wei Xu, Bingbing Song, Qun Ding, and Shu-Chuan Chu.
\newblock Prediction of chaotic time series of rbf neural network based on
  particle swarm optimization.
\newblock In \emph{Intelligent Data analysis and its Applications, Volume II},
  pages 489--497. Springer, 2014.

\bibitem[Dugas et~al.(2013)Dugas, Jalalpour, Gel, Levin, Torcaso, Igusa, and
  Rothman]{dugas2013influenza}
Andrea~Freyer Dugas, Mehdi Jalalpour, Yulia Gel, Scott Levin, Fred Torcaso,
  Takeru Igusa, and Richard~E Rothman.
\newblock Influenza forecasting with google flu trends.
\newblock \emph{PloS one}, 8\penalty0 (2):\penalty0 e56176, 2013.

\bibitem[Glorot and Bengio(2010)]{glorot2010understanding}
Xavier Glorot and Yoshua Bengio.
\newblock Understanding the difficulty of training deep feedforward neural
  networks.
\newblock In \emph{Proceedings of the thirteenth international conference on
  artificial intelligence and statistics}, pages 249--256, 2010.

\bibitem[Gong and Bowman(2018)]{gong2017ruminating}
Yichen Gong and Samuel Bowman.
\newblock Ruminating reader: Reasoning with gated multi-hop attention.
\newblock In \emph{Proceedings of the Workshop on Machine Reading for Question
  Answering}, pages 1--11, Melbourne, Australia, July 2018. ACL.
\newblock \doi{10.18653/v1/W18-2601}.

\bibitem[Hochreiter and Schmidhuber(1997)]{hochreiter1997long}
Sepp Hochreiter and J{\"u}rgen Schmidhuber.
\newblock Long short-term memory.
\newblock \emph{Neural computation}, 9\penalty0 (8):\penalty0 1735--1780, 1997.

\bibitem[Kermack and McKendrick(1927)]{kermack1927contribution}
William~Ogilvy Kermack and Anderson~G McKendrick.
\newblock A contribution to the mathematical theory of epidemics.
\newblock \emph{Proceedings of the royal society of london. Series A,
  Containing papers of a mathematical and physical character}, 115\penalty0
  (772):\penalty0 700--721, 1927.

\bibitem[Kim(2014)]{kim-2014-convolutional}
Yoon Kim.
\newblock Convolutional neural networks for sentence classification.
\newblock In \emph{Proceedings of the 2014 Conference on Empirical Methods in
  Natural Language Processing ({EMNLP})}, pages 1746--1751, Doha, Qatar,
  October 2014. ACL.
\newblock \doi{10.3115/v1/D14-1181}.

\bibitem[Kinga and Adam(2015)]{kinga2015method}
D~Kinga and J~Ba Adam.
\newblock A method for stochastic optimization.
\newblock In \emph{International Conference on Learning Representations},
  volume~5, 2015.

\bibitem[Kingma and Ba(2015)]{kingma2014adam}
Diederik~P. Kingma and Jimmy Ba.
\newblock Adam: A method for stochastic optimization.
\newblock In \emph{Proceedings of the 2015 International Conference on Learning
  Representations}, volume abs/1412.6980, 2015.

\bibitem[Kipf and Welling(2016)]{kipf2016semi}
Thomas~N Kipf and Max Welling.
\newblock Semi-supervised classification with graph convolutional networks.
\newblock \emph{arXiv preprint arXiv:1609.02907}, 2016.

\bibitem[Qian-Li et~al.(2008)Qian-Li, Qi-Lun, Hong, Tan-Wei, and
  Jiang-Wei]{qian2008multi}
Ma~Qian-Li, Zheng Qi-Lun, Peng Hong, Zhong Tan-Wei, and Qin Jiang-Wei.
\newblock Multi-step-prediction of chaotic time series based on co-evolutionary
  recurrent neural network.
\newblock \emph{Chinese Physics B}, 17\penalty0 (2):\penalty0 536, 2008.

\bibitem[Santos and Matos(2014)]{santos2014analysing}
Jos{\'e}~Carlos Santos and S{\'e}rgio Matos.
\newblock Analysing twitter and web queries for flu trend prediction.
\newblock \emph{Theoretical Biology and Medical Modelling}, 11\penalty0
  (1):\penalty0 S6, 2014.

\bibitem[Senanayake et~al.(2016)Senanayake, O'Callaghan, and
  Ramos]{senanayake2016predicting}
Ransalu Senanayake, Simon O'Callaghan, and Fabio Ramos.
\newblock Predicting spatio-temporal propagation of seasonal influenza using
  variational gaussian process regression.
\newblock In \emph{Thirtieth AAAI Conference on Artificial Intelligence}, 2016.

\bibitem[Sorjamaa et~al.(2007)Sorjamaa, Hao, Reyhani, Ji, and
  Lendasse]{sorjamaa2007methodology}
Antti Sorjamaa, Jin Hao, Nima Reyhani, Yongnan Ji, and Amaury Lendasse.
\newblock Methodology for long-term prediction of time series.
\newblock \emph{Neurocomputing}, 70\penalty0 (16-18):\penalty0 2861--2869,
  2007.

\bibitem[Sukhbaatar et~al.(2015)Sukhbaatar, Weston, Fergus,
  et~al.]{sukhbaatar2015end}
Sainbayar Sukhbaatar, Jason Weston, Rob Fergus, et~al.
\newblock End-to-end memory networks.
\newblock In \emph{Advances in neural information processing systems}, pages
  2440--2448, 2015.

\bibitem[Vaswani et~al.(2017)Vaswani, Shazeer, Parmar, Uszkoreit, Jones, Gomez,
  Kaiser, and Polosukhin]{vaswani2017attention}
Ashish Vaswani, Noam Shazeer, Niki Parmar, Jakob Uszkoreit, Llion Jones,
  Aidan~N Gomez, {\L}ukasz Kaiser, and Illia Polosukhin.
\newblock Attention is all you need.
\newblock In \emph{Advances in neural information processing systems}, pages
  5998--6008, 2017.

\bibitem[Venna et~al.(2018)Venna, Tavanaei, Gottumukkala, Raghavan, Maida, and
  Nichols]{venna2018novel}
Siva~R Venna, Amirhossein Tavanaei, Raju~N Gottumukkala, Vijay~V Raghavan,
  Anthony~S Maida, and Stephen Nichols.
\newblock A novel data-driven model for real-time influenza forecasting.
\newblock \emph{IEEE Access}, 7:\penalty0 7691--7701, 2018.

\bibitem[Viboud et~al.(2003)Viboud, Bo{\"e}lle, Carrat, Valleron, and
  Flahault]{viboud2003prediction}
C{\'e}cile Viboud, Pierre-Yves Bo{\"e}lle, Fabrice Carrat, Alain-Jacques
  Valleron, and Antoine Flahault.
\newblock Prediction of the spread of influenza epidemics by the method of
  analogues.
\newblock \emph{American Journal of Epidemiology}, 158\penalty0 (10):\penalty0
  996--1006, 2003.

\bibitem[Waller et~al.(1997)Waller, Carlin, Xia, and
  Gelfand]{waller1997hierarchical}
Lance~A Waller, Bradley~P Carlin, Hong Xia, and Alan~E Gelfand.
\newblock Hierarchical spatio-temporal mapping of disease rates.
\newblock \emph{Journal of the American Statistical association}, 92\penalty0
  (438):\penalty0 607--617, 1997.

\bibitem[Wang et~al.(2019)Wang, Chen, and Marathe]{Wang-2019}
Lijing Wang, Jiangzhuo Chen, and Madhav Marathe.
\newblock Defsi: Deep learning based epidemic forecasting with synthetic
  information.
\newblock volume~33, pages 9607--9612, Jul. 2019.
\newblock \doi{10.1609/aaai.v33i01.33019607}.

\bibitem[Wang et~al.(2015)Wang, Chakraborty, Mekaru, Brownstein, Ye, and
  Ramakrishnan]{wang2015dynamic}
Zheng Wang, Prithwish Chakraborty, Sumiko~R Mekaru, John~S Brownstein, Jieping
  Ye, and Naren Ramakrishnan.
\newblock Dynamic poisson autoregression for influenza-like-illness case count
  prediction.
\newblock In \emph{Proceedings of the 21th ACM SIGKDD International Conference
  on Knowledge Discovery and Data Mining}, pages 1285--1294. ACM, 2015.

\bibitem[Wu et~al.(2018)Wu, Yang, Nishiura, and Saitoh]{wu2018deep}
Yuexin Wu, Yiming Yang, Hiroshi Nishiura, and Masaya Saitoh.
\newblock Deep learning for epidemiological predictions.
\newblock In \emph{The 41st International ACM SIGIR Conference on Research \&
  Development in Information Retrieval}, pages 1085--1088. ACM, 2018.

\bibitem[Yang et~al.(2014)Yang, Karspeck, and Shaman]{yang2014comparison}
Wan Yang, Alicia Karspeck, and Jeffrey Shaman.
\newblock Comparison of filtering methods for the modeling and retrospective
  forecasting of influenza epidemics.
\newblock \emph{PLoS computational biology}, 10\penalty0 (4):\penalty0
  e1003583, 2014.

\end{thebibliography}
